\documentclass[10pt,journal,compsoc]{IEEEtran}
\ifCLASSOPTIONcompsoc
\usepackage[nocompress]{cite}
\else
\usepackage{cite}
\fi

\usepackage{graphicx}

\usepackage{float}

\usepackage{tabularx}

\usepackage{makecell}

\usepackage{multicol}

\usepackage{booktabs}
\usepackage{multirow}

\usepackage{amssymb}
\usepackage{pifont}

\usepackage[T2A,T1]{fontenc}
\usepackage[utf8]{inputenc}
\usepackage[russian,english]{babel}

\ifCLASSINFOpdf
\else
\fi
\begin{document}

		%
		\title{Large language models for artificial general intelligence (AGI): A survey of foundational principles and approaches}
		%
		%
		%
		%

		\author{Alhassan~Mumuni$^1$$^*$~\IEEEmembership{}
			and~Fuseini~Mumuni$^2$~\IEEEmembership{}
			\IEEEcompsocitemizethanks{\IEEEcompsocthanksitem $^1$Alhassan Mumuni:
				Department
				of Electrical and Electronics Engineering, Cape Coast Technical University, Cape Coast, Ghana.\protect\\
				$^*$Corresponding author, E-mail: alhassan.mumuni@cctu.edu.gh
				\IEEEcompsocthanksitem $^2$Fuseini Mumuni: University of Mines and Technology, UMaT, Tarkwa, Ghana. E-mail: fmumuni@umat.edu.gh}
		}

		%
		%

	\markboth{}%
	{Shell \MakeLowercase{\textit{et al.}}: Bare Demo of IEEEtran.cls for Computer Society Journals}
	%



	\IEEEtitleabstractindextext{%
		\begin{abstract}

	Generative artificial intelligence (AI) systems based on large-scale pretrained foundation models (PFMs) such as vision-language models, large language models (LLMs), diffusion models and vision-language-action (VLA) models have demonstrated the ability to solve complex and truly non-trivial AI problems in a wide variety of domains and contexts. Multimodal large language models (MLLMs), in particular, learn from vast and diverse data sources, allowing rich and nuanced representations of the world and, thereby, providing extensive capabilities, including the ability to reason, engage in meaningful dialog; collaborate with humans and other agents to jointly solve complex problems; and understand social and emotional aspects of humans. Despite this impressive feat, the cognitive abilities of state-of-the-art LLMs trained on large-scale datasets are still superficial and brittle. Consequently, generic LLMs are severely limited in their generalist capabilities.  A number of foundational problems —embodiment, symbol grounding, causality and memory — are required to be addressed for LLMs to attain human-level general intelligence. These concepts are more aligned with human cognition and provide LLMs with inherent human-like cognitive properties that support the realization of physically-plausible, semantically meaningful, flexible and more generalizable knowledge and intelligence. In this work, we discuss the aforementioned foundational issues and survey state-of-the art approaches for implementing these concepts in LLMs. Specifically, we discuss how the principles of embodiment, symbol grounding, causality and memory can be leveraged toward the attainment of artificial general intelligence (AGI) in an organic manner.  
			
		\end{abstract}
		
		}

	\maketitle

	\IEEEdisplaynontitleabstractindextext

	%
	\IEEEpeerreviewmaketitle

	\section{Introduction}

	\subsection{Background}
	
	Intelligence relates to the ability of a system, biological or otherwise, to achieve some level of success in accomplishing one or more desired goals in a given environment (or variety of environments). An intelligent system is capable of inferring its own state as well as the state of the environment, and is able to transform these inferences into appropriate responses leading to the achievement of desired goals. Intelligence is characteristically a unique feature of higher living organisms, and in the pursuit of developing their artificial counterparts, artificial intelligence, researchers have frequently borrowed concepts from biology. An important attribute of biological intelligence is its generality, i.e., its ability to handle many different problems across a wide variety of settings. Human intelligence, in particular, is remarkably sophisticated, rich and versatile, and can effortlessly handle many novel tasks. The general superiority of human intelligence over that of other higher animals stems (primarily) from the ability of humans to structure and transfer knowledge through social and cultural constructs such as art, norms, rituals, belief systems and customs \cite{adams2012beyond}. Language plays a vital role in all these processes.

	While the idea of creating this kind of generalist intelligence is attractive, it is extremely challenging to achieve such level of sophistication and generalization power in machines. Until quite recently, AI techniques that achieved impressive results were narrowly focused, solving specific problems in one domain or in a restricted set of domains (e.g., face recognition, medical image segmentation, text translation, stock market forecasting, pedestrian tracking, etc.). Lately, generative AI techniques based on variational autoencoders (VAEs) \cite{kingma2013auto} and generative adversarial networks (GANs) \cite{isola2017image} have contributed greatly in revolutionizing the capabilities of AI, and enabling single models to simultaneously handle a wide variety of complex tasks \cite{bond2021deep}. More recently, the emergence of large-scale pretrained foundation models such as large language models (LLMs) \cite{minaee2024large}, diffusion models (DMs) \cite{croitoru2023diffusion}, vision-language models (VLMs) \cite{du2022survey} and vision-language-action (VLA) models \cite{ma2024survey} has real prospect for replicating generalist property in artificial intelligence.  Owing to their ability to handle a wide range of challenging open-domain problems \cite{bubeck2023sparks,y2023artificial,firoozi2023foundation,yu2024seqgpt }, large-scale pretrained foundation models, especially multimodal large language models, have renewed interest in the quest for developing artificial general intelligence \cite{y2023artificial}. The main aim of this work is to present the fundamental principles of cognition that supports the realization of artificial general intelligence, and review state-of-the-art techniques for implementing these concepts in large language models.

	\subsection{Language as the foundation of general intelligence in biological systems}

	\subsubsection{Language as a medium of knowledge acquisition, representation and organization}
	It has been shown that communication using natural language is one of the most effective ways of learning general knowledge about the real world \cite{mohanty2023transforming}, and while, human's sensory and motor capabilities are not generally superior to other higher animals, including primates (see \cite{spelke2003makes,prestrude1970sensory,doble2014animal,mendoza2014motor,felleman1991distributed,thelen1984learning,ackermann2014brain}, human cognitive capabilities are far more advanced than other animals. The superiority of man’s cognitive capacity compared to other members of the animal kingdom, including his closest relative, primates, has been largely attributed to humans’ use of language \cite{novack2020becoming,bourguignon2023emergence, grigorenko2023never}.

	Language plays a central role in man’s ability to represent, interpret and reason with abstract concepts \cite{clark2012magic}. In human societies, one of the most important functions of language is to facilitates the acquisition and sharing of new knowledge. With the help of language – whether by literature, speech or art – humans can effortlessly learn from others and accumulate knowledge not only by observation or through their own interactions with the world, but also by acquiring knowledge accumulated by other humans. Besides, language provides a conceptual framework for representing and internalizing knowledge \cite{bourguignon2023emergence}. It has been demonstrated that the specific linguistic structures and vocabulary used by a group influence reasoning and interpretation of the world. Indeed, linguistic differences (e.g., in terms of vocabulary) has been shown to influence how individuals members of different linguistic groups remember and describe their experiences \cite{schmitt1994language,tajima2012linguistic,trueswell2010perceiving,feist2013remembering}. In this regard, language can structure or restructure cognition \cite{majid2004can}, and therefore shapes how subjects understand and interact with the world \cite{duranti2011linguistic,lupyan2016centrality}.

	\subsubsection{Language as a tool for cognitive information processing} 
	
	Besides creating abstractions to represent and organize the representation of perceptual information and knowledge, language plays a fundamental role in facilitating cognitive computational operations \cite{clark2012magic}. Lupyan \cite{lupyan2016centrality} argues that basic linguistic elements like words provide cues for other cognitive components to construct meaning. Thus, language is not just a set of static symbols that reference real-world objects, phenomena and experiences, but it also serves as a tool for manipulating these symbols. Clark \cite{clark2012magic} specifically describes six different ways by which language facilitates cognitive information processing and reasoning in humans. Language been shown to facilitate not just crystalized intelligence (i.e., representation-related cognitive mechanisms) such as experience/stimuli categorization \cite{tajima2012linguistic} and memory \cite{schmitt1994language,feist2013remembering} but also elements of fluid intelligence (i.e., analytical problem-solving skills) like perception \cite{thierry2009unconscious,lupyan2008perceptual,meteyard2007motion} and reasoning \cite{clark2012magic,lupyan2016centrality}. Moreover, exposure to multiple linguistic frameworks has been demonstrated to broaden the individual’s perspective and facilitates an understanding of concepts in a more nuanced manner. Because of its centrality in biological cognitive abilities, language has been characterized variously as “the interface to cognition” \cite{novack2020becoming}, “intelligence amplifier” \cite{roth2012evolution}, and human cognition itself has been described as language-augmented cognition \cite{lupyan2016centrality}.

	\subsection{The concept of artificial general intelligence}
	
	While there are different interpretations of artificial general intelligence (AGI) in the literature \cite{goertzel2014artificial,mclean2023risks,wang2007introduction,fei2022towards,bubeck2023sparks,ilic2024evidence}, the concept is generally understood as AI systems that exhibit broad intellectual abilities and are able to perform high-level cognitive tasks such as perception – including context understanding and a degree of self-awareness \cite{aliman2018hybrid,buttazzo2023rise}, reasoning, planning, and the application of learned knowledge in new contexts.  AGI systems are universally powerful models that can successfully accomplish significantly complex and diverse cognitive tasks across multiple domains without the need for additional training. The term human-level intelligence \cite{adams2012mapping,besold2016generality,mclean2023risks} is often loosely used to refer to AI systems that demonstrate general intelligence.

\begin{figure*}[!htb]
		
		\vspace {-6mm}
		\centering
		\includegraphics[width=0.80 \linewidth]{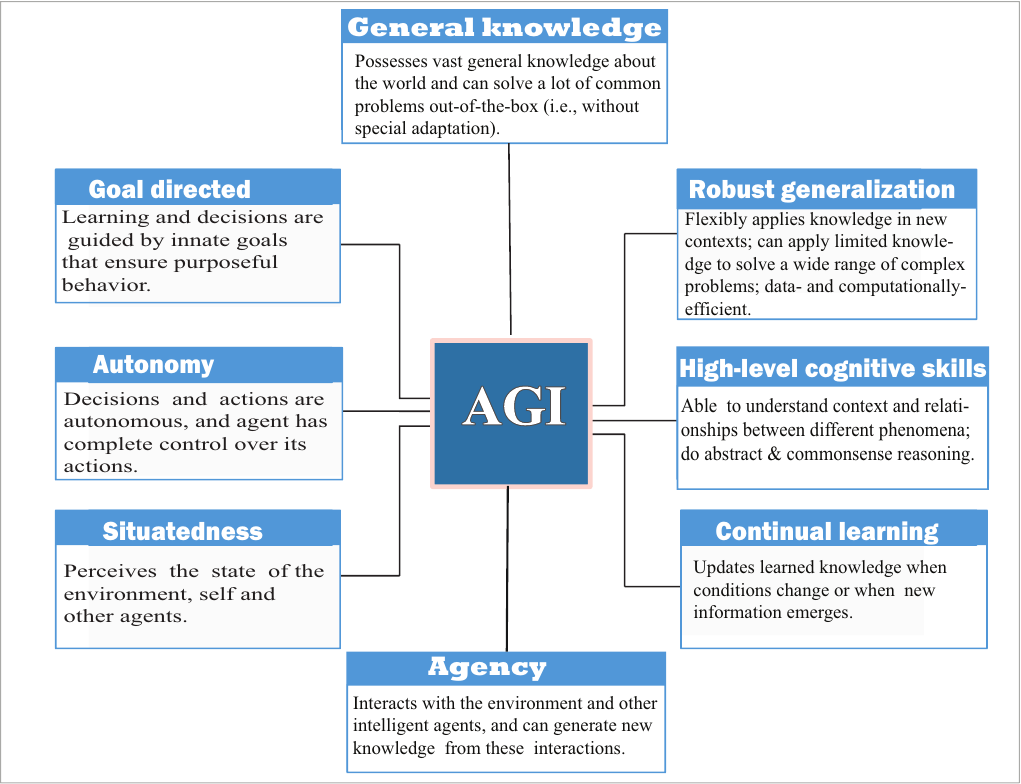} \vspace {-3mm}
		\caption{Some of the most important features of artificial general intelligence (AGI) systems. These features give AGI systems vast cognitive capabilities despite the models' limited knowledge and the need to, for the sake of conserving energy and time, take shortcuts in cognitive information processing.  
		}\label{fig:1_AGI_Properties}
	\end{figure*}

	AGI should not be taken to mean super-omniscience and omnipotent machines. Such hypothetical level of capability is referred to as artificial super-intelligence \cite{pueyo2018growth,bostrom2020ethical}. Practical AGI systems are systems possessing general – yet limited and, to a degree, uncertain – knowledge about the world but is sufficiently powerful and flexible to solve a wide range of problems requiring sensorimotor control, perception, context understanding, commonsense and analytical reasoning capabilities. This understanding of artificial general intelligence, in essence, reflects not only the fact of the practical difficulties in embedding or learning all relevant knowledge and skills at once, but also the performance limitations of such an approach. Moreover, conceptualizing artificial general intelligence as limited in scope but adaptive, flexible and extensible is consistent with the nature and properties of biological intelligence in higher living organisms like humans. Despite the wide variety of definitions in the literature, there almost a unanimous agreement on some of the defining features of AGI. Specifically, the most important features of a typical AGI system are that (see, for example, \cite{goertzel2014artificial,bubeck2023sparks,adams2012mapping,pennachin2007contemporary,voss2007essentials}): it can learn and flexibly apply the limited and uncertain knowledge to solve a wide range of problems in entirely different contexts; its learning and actions are autonomous and goal-driven; it retains and accumulates relevant information in memory and reuse the knowledge in future tasks; and it can understand context and perform high-level cognitive tasks such as abstract and commonsense reasoning. We summarized the important properties in Figure \ref{fig:1_AGI_Properties}.


	It is important to point out that AGI is fundamentally different from Strong AI (see \cite{braga2017emperor,flowers2019strong,butz2021towards}). While AGI focuses on developing intelligent systems that have broad cognitive capabilities and can solve truly nontrivial problems, Strong AI aims to create very powerful intelligence that not only mimics human cognitive abilities at the functional level but one that is also characterized by real human cognitive properties such as intrinsic mental states and subjective experiences, including intentionality (desires, hopes, beliefs, inner motivation, etc.), morality, emotions, and self-awareness \cite{mograbi2024cognitive,takeno2012creation} in the sense of being conscious and sentient. Readers interested in this topic may want to see \cite{metz2022ai,knight2019refuting,gibert2022search,ng2020strong,mcdermott2007artificial} for more detailed discussions on Strong AI concepts, including sentience \cite{metz2022ai,gibert2022search,ng2020strong}, consciousness \cite{knight2019refuting,ng2020strong,li2021ai} and morality \cite{manna2021problem,young2019autonomous} of AI systems.

	\subsection{Scope and outline of work}

	In this work, we present an extensive discussion of the core principles we consider important to achieving general intelligence. We also discuss the various approaches for realizing each of these concepts in artificial intelligence and LLM systems. The concepts discussed here are not algorithmic solutions for achieving AGI but rather general principles and properties of biological intelligence that AI systems based on large language models must be imbued with if they are to attain the kind of generality, robustness and sophistication of human cognitive functions. In fact, the core concepts are by nature algorithm-agnostic, that is, their implementation is not specific to any particular techniques or set of methods. It is important, however, to note that specific cognitive functions – e.g., perception, reasoning, planning, action, etc. – can be enriched by these general concepts and principles.  
	The remainder of the paper is organized as follows. In section 2, we present a brief overview of the key elements of large language models that make then so powerful and underlie their potential for solving complex problems requiring human-level general intelligence.  The important foundational principles for achieving general intelligence in large language models are covered in sections 3 through 6. These include embodiment (Section 3), symbol grounding (Section 4), causality (Section 5) and memory (Section 6).  In Section 7, we discuss the interrelationships and interactions of the cognitive principles and synthesize a holistic cognitive model based on these interrelationships and interactions.  Finally, we present a summary discussion of the concepts in Section 8 and conclude in Section 9.

	\section{Towards artificial general intelligence with large language models}
	\subsection{Large language models and artificial general intelligence}
	Much of human knowledge and skills have been acquired and transmitted through multiple media, most significantly through language and visual media (reading, listening, direct observation, etc.). In a similar way, multimodal language models, relying on multiple data modalities, hold a great promise for providing systems with general, multidimensional knowledge about the world. While unimodal language models such as GPT-3 \cite{floridi2020gpt} and BERT \cite{kenton2019bert} could handle only text data, multimodal LLMs (e.g., Palm-E \cite{driess2023palm}, Minigpt-4 \cite{zhu2023minigpt}, Flamingo \cite{alayrac2022flamingo}, LLaVA \cite{liu2024visual}) naturally integrates many different data modalities, including visual, auditory, textual and spatial information seamlessly to generate richer and more comprehensive representations for cognitive tasks. This is similar to the way biological intelligence relies on complex, multisensory data streams. 
	The generalist capabilities of state-of-the-art multimodal large language models have already been widely demonstrated \cite{bubeck2023sparks,ilic2024evidence}, and their ability to solve a wide range of complex cognitive problems that traditionally required human intelligence is in no doubt. The remarkable success of large language models has redefined the possibilities and scope of artificial intelligence. The main factor that drives this success is the ability to build and train very large neural network models on diverse, multimodal data. These models are typically trained on generic data from the wild (e.g., online publications, books, news articles, social media and other sources of information from the web), and are able to capture intricate concepts and generalize more effectively to new tasks with little (few-shot learning \cite{gao2020making,liu2023large}) or no (zero-shot learning \cite{kojima2022large}) task-specific training. Consequently, complex cognition-intensive, open-domain tasks such as commonsense and analytical reasoning \cite{yin2024survey,krause2024data}; mathematical problem-solving \cite{ahn2024large,imani2023mathprompter}; itinerary planning \cite{tang2024synergizing,kanhed2024destinai} or general task planning \cite{shen2024hugginggpt}; and open-vocabulary question answering \cite{cheng2024efficienteqa,majumdar2024openeqa}. Significantly, state-of-the-art LLMs are able to perform creative and artistic works such as composing essays, short stories, or even entire novels \cite{yuan2022wordcraft,franceschelli2024creativity} according to any given criteria (e.g., author style, diction, mood, etc.).


	\begin{figure*}[!htb]
		
		\vspace {-6mm}
		\centering
		\includegraphics[width=0.80 \linewidth]{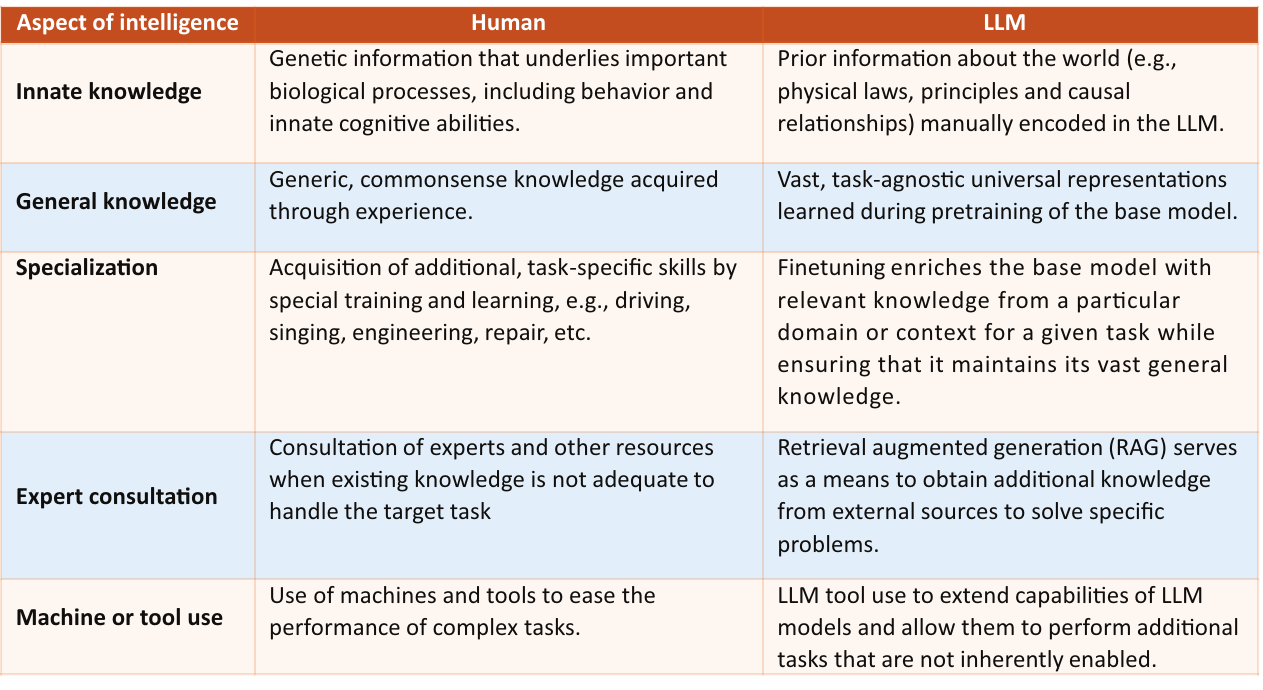} \vspace {-3mm}
		\caption{LLM versus human intelligence: Important mechanisms that allow flexible extension of knowledge and cognitive abilities.  
		}\label{fig:2_AGI_Human}
	\end{figure*}
	
	
	\subsection{Features of large language models that support the attainment of human-level intelligence}
	In the context of achieving general intelligence, besides training on large and diverse datasets, large language models possess a number of interesting features that allow their knowledge and skills to be naturally extended as needed. This extensibility, together with their already vast generic knowledge, allows them to overwhelmingly outperform traditional deep learning models that are typically designed with narrow optimization objectives and trained on restricted datasets from curated environments.  
	
	While the underlying processes and mechanisms that support extensibility of large language models are fundamentally different from those that support biological intelligence, the resulting properties somehow mirror the multilayered and multidimensional nature of human intelligence in many respects. For example, pretraining large language models endows them with general knowledge that is sufficiently powerful and flexible to tackle a wide range of common problems requiring perception, context understanding as well as commonsense and analytical reasoning capabilities. Where domain-specific knowledge is needed, finetuning can be applied to augment the general knowledge with specialized knowledge by training the pretrained LMM further on domain-specific datasets. This approach is similar to the way human experts – who already have general or commonsense knowledge– acquire specialized competencies in narrow areas of endeavor (e.g., as professionals in engineering, medicine, law, or web development). It is also usual to ground the internal representations in real-world concepts using prior knowledge. Again, this feature is similar to the way biological intelligence is built on prior knowledge encoded as genetic information. In addition to the internalized knowledge and cognitive capabilities, humans frequently rely external knowledge (e.g., through consultations with experts or books) and tools (e.g., software, machines, etc.) to extend their capabilities. Similarly, state-of-the-art language models can utilize tools (see \cite{shen2024llm,li2024review,wang2024tool}) and external knowledge –through retrieval augmented generation (RAG) \cite{gao2023retrieval,sanmartin2024kg} – to extend their capabilities. We summarized these important features in Figure \ref{fig:2_AGI_Human}.


	\begin{figure*}[!htb]
		\vspace {1mm}
		\centering
		\includegraphics[width=0.80 \linewidth]{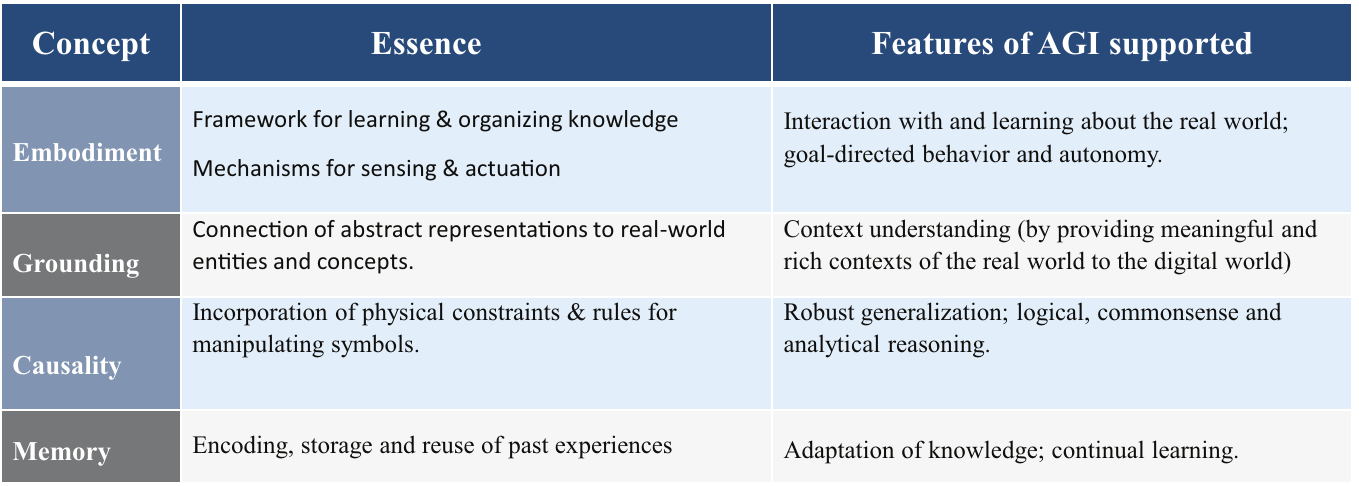} \vspace {-3mm}
		\caption{A summary of the essence and role of each of the foundational AGI concepts covered in this work.  
		}\label{fig:3_AGI_Role}
	\end{figure*}
	
	
	\subsection{Overview of foundational principles for AGI with LLMs}
	Despite the fact that state-of-the-art large language models are incredibly powerful, they still have a number of limitations that constrain their ability to achieve general intelligence \cite{ye2023cognitive,wolf2023fundamental,tamkin2021understanding,hadi2023survey}. Generally, the models’ understanding of context is often superficial and their solutions, in many cases, only exhibit external resemblance to human knowledge \cite{zevcevic2023causal,bender2021dangers}. The problem stems from the fact that AI systems, including LLMs, are still just digital constructs that that attempt to mimic human knowledge and cognitive capabilities by learning general properties of the world from vast amounts of data. This knowledge is generally limited to observed patterns but does not capture the underlying principles responsible for the behavior. 
	
	It has long been argued that for machines to achieve AGI they necessarily need to emulate some key aspects of human cognition which enables human intelligence to be so robust, efficient, flexible and general yet sophisticated in the way it handles complex problems. Among the key aspects of human cognitive process are embodied sentience or simply embodiment \cite{howard2019evolving,durt2023against}, symbolic grounding \cite{harnad1990symbol}, causal reasoning \cite{goddu2024development,marwala2015causality}, and memory \cite{schweizer2004attention,conway2003working,alexander1997intelligence}. Embodied sentience—the ability to have subjective experiences and feel sensations—is considered a fundamental aspect of higher intelligence. It is an essential capability to enable general intelligence because it provides a sort of pseudo-consciousness and autonomy \cite{torres2024embodied,lin2024embodied,chalmers2023could} Specifically, it enables agents to be self-aware, and therefore align their decisions and actions to a more universal, intrinsic higher-level goal \cite{pezzulo2014principles,morsella2020motor,barandiaran2024transforming}. Embodied sentience also allows agents to recognize the experiences of others. This allows them to be ethical and moral in decision-making and behavior. Another key principle of biological intelligence, symbolic grounding, performs a complementary function to embodiment by connecting abstract cognitive representations to meaningful entities and concepts in the real world.  Grounding in LLMs ensures that the abstract representations learned correspond to specific real-world concepts, and are utilized or manipulated within the context of their semantic essence. Although the internal mechanisms underlying the grounding process in human cognition are still not well-understood, rudimentary techniques for realizing grounding in artificial intelligence systems have shown a lot of promise in their ability to align
	LLMs’ knowledge with the world’s \cite{carta2023grounding}. Another important set of ingredients for artificial general intelligence, intuitive physics \cite{vicovaro2021intuitive,duan2022survey} and intuitive psychology \cite{baron2001intuitive,ross1977intuitive}, relate to the ability to infer cause-and-effect relations about events and interactions in the real world. Human’s natural understanding of intuitive physics is known to be the basis of robust perception and causal reasoning abilities. Meanwhile, intuitive psychology allows humans to form beliefs about intentions and probable actions of other living entities without the need to learn about specifically learn about them. Theory of mind (ToM) techniques \cite{verma2024theory,li2023theory,richards2024you} are typically employed in LLMs to facilitate their understanding of intuitive psychology.  Finally, memory allows learned knowledge and past experiences to be preserved and accumulated over time. This extends and enriches knowledge in a way that promotes general-purpose utility. Moreover, the ability to introspect \cite{toy2024metacognition,qu2024recursive} and reflect \cite{yao2023retroformer} on past decisions and actions by virtue of memory mechanisms provides a way for LLMs to (accomplish) continual learning and adaptation. A summary of the role of each of these concepts is presented in Figure \ref{fig:3_AGI_Role}.

	\section{Embodiment}
	
	\subsection{Basic concept of embodiment}
	Modern conceptualizations of biological cognition suggest that cognitive processes in the human nervous system are deeply rooted in the mind’s interactions with the body and the external environment. Per this view of intelligence, the brain, body, and environment are assumed to form a unified system where they jointly influence and shape intelligent behavior \cite{mougenot2024theoretical,wilson2013embodied,tibbetts2014does,foglia2013embodied}). The concept of neural plasticity, one of the most important cognitive phenomena that enhance adaptative behavior of intelligence, also assumes mind-body-environment interaction \cite{hsu2024entangled,thoenen1995neurotrophins,huttenlocher2009neural}. In \cite{wilson2013embodied}, the three components are considered as essential cognitive resources that are required by the organism to solve specific tasks. 
	In line with this understanding of mind-body-environment trinity, it has been argued that for AI systems to be truly intelligent, they, like biological systems, must necessarily be able to interact with the world in a physical way and receive feedback and learn about the results of those physical interactions (see \cite{pfeifer2004embodied,sharkey2001mechanistic,savva2019habitat,mecattaf2024little,liu2024aligning,zhang2023large}). According to this hypothesis, artificial intelligence can only attain general intelligence comparable to human-level cognitive capabilities if such intelligence were created in, and intrinsically linked with a physical body that possesses the ability to perform physical actions on the environment\cite{liu2024aligning,duan2022survey,lee2006physically}. Embodied AI systems are systems that – unlike traditional approaches that are solely digital in nature – have a tangible physical manifestation through which they can perceive and process sensory information, and interact with their environment. 
	
	\subsection{Embodiment as the foundation of general intelligence}
	Embodiment provides the foundation for intrinsic goal-directed behavior. An embodied artificial intelligence system necessarily has agency \cite{swanepoel2021does,aagerfalk2020artificial}, i.e., it undertakes intentional actions – actions it desires to perform (e.g., based on specific goals and needs) and over which it has complete control. This goal-directed behavior is a fundamental requirement for autonomy. Besides, AI systems endowed with rich sensorimotor resources with unlimited possibilities to explore and interact with the environment will attain extensive intellectual capabilities.  Such an intelligent system will necessarily possess accurate and robust perception of the world and of its own state. In addition, it must be able to act on and influence the world in a purposeful way. It must also be capable of seamlessly adapting to the complex dynamics of the real world.  Thus, while conventional approaches to intelligence results in models that are inherently rigid and mechanistic, embodied intelligence is more flexible and nuanced, and can connect objective experiences with subjective concepts values, cultural norms and expectations \cite{williams2022belief}.  By providing a more integrated way of interacting with the world, learning and decision-making, embodied agents are more robust and can handle complex and diverse problems, thereby supporting their generalist credentials.

	\subsection{Key aspects of embodied intelligence}
	The most important aspect of artificial general intelligence is the requirement for full autonomy – the ability to independently make decisions and take appropriate actions even in the absence of explicit commands or control signals from the outside world.  The implementation of embodied general intelligence involves four main considerations. 
	
	\textbf{Goal-awareness:} In order to achieve full autonomy, like biological systems, the artificial intelligence system must have an overarching goal to which all other goals, including explicit instructions given by other actors, must be subordinated. This goal must be intrinsic and guide the successful accomplishment of external goals triggered by other agents (e.g., commands given by users, actions of other agents, etc.).
	
	\textbf{Self-awareness:} As the body is the executor of actions that influence the physical world, the intelligence process must be tied to the structure and capabilities of the body. That is, the appropriateness of intelligent actions depends not only on the goal the intelligent system seeks to achieve, but also on the optimality of the actions with respect to the available means to carry out the target goal. In order to be successful, the embodied intelligent agent must, therefore, be aware of its own capabilities and limitations. In human-centric contexts, this awareness includes the ability to understand oneself from the perspective the broader social setting, and to connect experiences with values, cultural norms and expectations. This facilitates the realization of social intelligence \cite{williams2022belief,lindblom2015embodied,voestermans2013culture}.
	
	\textbf{Situational-awareness:} An entity's intelligence is shaped by the specific context or situation it finds itself in. This underscores the fact that intelligent behavior is often a response to specific needs or challenges presented by the environment. Therefore, to achieve any goal, it is important to know the properties of the world and to predict beforehand the outcome of the target action with respect to the intended goal of the action. Moreover, since the external world behaves differently in response to actions by different entities, the intelligence of each intelligent system must be unique in some way. This means intelligent behavior of the AI system must take into account the expected responses elicited by other objects or the environment by virtue of its special characteristics. In humans, behavior is often shaped by social and cultural factors. Similarly, the actions of embodied AI agents must reflect social, cultural and demographic realities of their environment. AI agents must be able to achieve goals while respecting practical constraints, including safety \cite{gao2024coca} and alignment with ethics and cultural values \cite{agarwal2024ethical}.
	
	\textbf{Deliberate action:} Actions are central to embodied intelligence since they are the primary means to influence the world and to achieve desired goals. Through actions an agent can perform active exploration of the world, thus further improving its perception and facilitating learning and adaptation in dynamic environments. Intelligent embodied agents must incorporate mechanisms to influence the world trough purposeful actions.

	In the following subsections, we discuss these four aspects of embodied intelligence and approaches for realization with modern AI systems based on pretrained foundation models, particularly large language models. A detailed summary of these discussions is presented in Figure \ref{fig:8_All_Summary_Emb}

	\subsection{Goal-awareness}
	
	\subsubsection{Foundation of goal-awareness and role in general intelligence}
	Human behavior is generally guided by goals that extend far beyond the objectives of immediate tasks \cite{ajzen1986prediction,frese2021goal,aarts2000habits}. These high-level goals are an important aspect of biological intelligence \cite{duncan1996intelligence}. Similarly, in machines, intelligence is intrinsically linked with the ability to achieve defined goals. Therefore, to achieve truly general intelligence in AI, that kind of high-level goal-oriented behavior is required. Goal-awareness is considered a crucial capability for the realization of artificial general intelligence because it determines the ability of AI systems to operate autonomously \cite{morrisposition,hansmeier2023xcs,goertzel2014golem}. Specifically, high-level goals provide intrinsic guidance that ensures meaningful and purposeful behavior in the absence of an external influence or instructions. 
	
\begin{figure}[H]
		\vspace {-1mm}
		\centering
		\includegraphics[width=1.0 \linewidth]{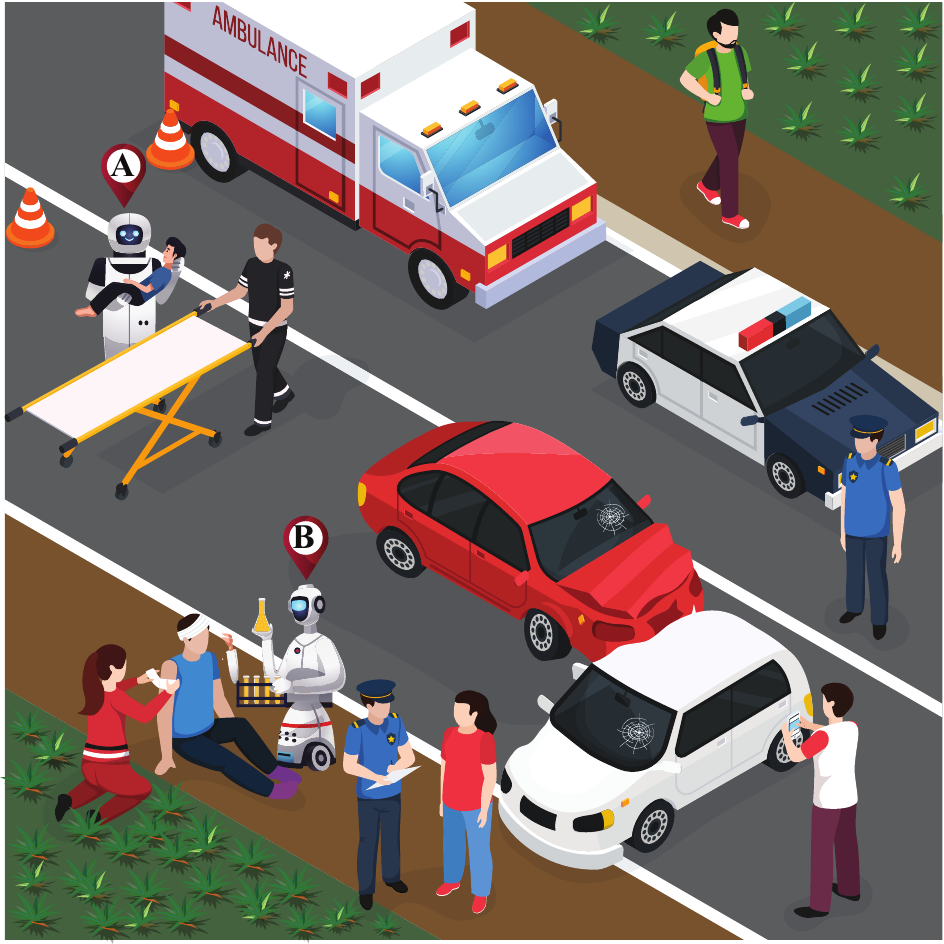} \vspace {-3mm}
		\caption{In this scene, two intelligent agents \textbf{A} and \textbf{B} assist during an emergency. When driven by high-level goals that are aligned with human interests and values, such agents can perform good acts spontaneously. Goal-awareness allows them to be proactive, autonomous and capable of attending to multiple tasks without deviating from their main essence. 
		}\label{fig:63E_Awareness_Goal}
	\end{figure}

	Goal-driven embodied agents can align immediate decisions and actions towards useful, long-term outcomes. In this regard, goal-oriented behavior facilitates an open-ended approach to problem-solving, allowing intelligent agents to exploit many possible actions without being restricted to specific behavioral options. This flexibility is important in problem settings where the course of action is not immediately obvious or cannot be computed analytically or is ill-defined and require non-linear, creative reasoning to arrive at. In particular, it allows autonomous AI agents to perform useful acts in society, for example, responding to emergencies like motor accidents, while still maintaining their core functions. In Figure \ref{fig:63E_Awareness_Goal}, for instance, two intelligent agents are shown taking part in evacuating and assisting victims during a traffic accident. These agents may not have been trained for such specific acts and may not have even anticipated such an incident but, being directed a higher goal that is aligned with broader societal values, they can make decisions independently to assist in such situations. Besides this cognitive flexibility, AI systems with goal-awareness capability can better generalize learning, select or prioritize relevant knowledge, and pursue actions that are directed towards achieving specific outcomes.  
	
	Goal-driven behavior is particularly important in long-horizon tasks, situations involving delayed reward \cite{hasselmo2005model,kopetz2021motivational,al2018complexity}, where immediate actions do not have a direct correlation with current sensory state of the agent. In this case, the intelligent agent selects actions based on intrinsic goals rather than explicit instructions. Many intelligent tasks performed by animals involve such delayed rewards \cite{hasselmo2005model,luo2022diffusion}. The goals in biological cognition can be in different forms and from different sources \cite{hommel2022goaliath}, including implicit goals driven by biological needs (e.g., survival, reproduction, etc.), persistent, temporary objectives for a particular task (mission),  instructions given by other humans.

	\subsubsection{Approaches to achieving goal-awareness in LLMs}
	Generic LLMs have been shown to exhibit goal-oriented behavior \cite{yang2024selfgoal,bellos2024can,ruis2023llms}. Notwithstanding these recent capabilities, goal-awareness in LLMs out of the box is still limited. For instance, experimental evaluation of LLM goal-awareness capabilities by Li Yu et al. \cite{li2024think}and Li Chuang et al. \cite{li2024incorporating} show poor goal-awareness. To mitigate this shortcoming, some recent works (e.g., \cite{deng2023goal,deng2023rethinking,ni2022hitkg}) have sought to align LLM behavior with explicitly-specified goals. One of the simplest ways to introduce goal-awareness in LLMs is to incorporate high-level goals in the form of input prompts for the LLM to guide the underlying models \cite{li2024towards,valmeekam2022large,yao2023value}. Approaches to enabling goal-oriented behavior in a more intrinsic manner involve specifically formulating goals in the LLM framework \cite{zhong2021keyword,deng2023unified}. For instance, Li et al. \cite{li2024incorporating} employ a dedicated goal planning agent together with a tool-augmented knowledge retrieval agent to handle goal-awareness in long-horizon tasks.  Liu et al. \cite{liu2022graph} encoded goal information in a knowledge graph which is then leveraged to design a goal planning module that guides LMM-human conversations in a goal-directed manner. Similarly, Ni et al. \cite{ni2022hitkg} exploit the commonsense relationships that exist in knowledge graph entities as goals for conversational LLM agents. With the approach, goal-directed responses are generated by traversing through the graph. Another common approach is to finetune the LLM on specific datasets (e.g., \cite{perez2022discovering,zhou2018towards,li2024think,sarkar2024gomaa}) that have been curated with the intended goal in mind. Unfortunately, models trained this way are often short-term goal-oriented. Another way to enhance long-term goal-awareness can also be accomplished by fine-tuning LLMs with imitation learning (e.g., in \cite{snell2022context}), using reinforcement Learning with Human Feedback \cite{volovikova2024instruction} or with feedback from a submodule (e.g., \cite{qin2024mp5}) or from a different LLM acting as external evaluator (e.g., \cite{barj2024reinforcement}). Advanced LLMs can leverage intrinsic high-level goal-awareness to enable intelligent agents to independently formulate low-level goals and pursue task-specific objectives without explicit human supervision have already been proposed \cite{yang2024selfgoal}. In EmbodiedGPT \cite{mu2024embodiedgpt} and CoTDiffusion \cite{ni2024generate}, for instance, Chain of Thought approach is used to generate subgoals for embodied actions. They can also refine decisions and modify actions according to changing circumstances and goals \cite{yang2024selfgoal,yuan2024pre}. 
	
	\subsubsection{Scope of application of goal-awareness in intelligent agents}
	Goal-awareness can facilitate human-robot collaboration \cite{chen2017guest,deng2023goal,chen2012situation}. When LLM-based intelligent virtual agents or robots are aware of goals—both their own and those of the humans they work with—they can align their actions more closely with human intentions, leading to more seamless and effective collaboration. Agents can then be more forward-looking and take proactive actions \cite{deng2023goal} instead of merely responding to user requests. The awareness of human goals also helps LLM agents to clarify ambiguous situations and better interpret observations about humans.  For instance, with knowledge of the broader goals, especially recommender-based conversational LLM agents, can provide better and more tailored responses \cite{wang2024improving}.  Moreover, high-level goals can provide context for understanding instructions and other human inputs.
	
	\subsubsection{Global and local goals}
	While biological cognition can handle global, high-level goals, LLMs, till date, are generally limited to tasks that can be described by or decomposed into multiple subtasks that, each consisting of a sequence of steps, where some kind of fixed ordering of actions exists. This kind of goal-awareness can be more accurately described as mission-awareness. In complex, real-world scenarios, intelligent agents need to understand not just immediate goals, or missions, but also how multiple intermediate – often seemingly contradictory – goals and subgoals fit into broader contexts, including societal interests (e.g., avoiding physical harm, minimizing climate change or promoting inclusiveness). Ultimately, the ability to incorporate/understand high-level goals allow AGI to reason about trade-offs and determine the best courses of action to maximize overall success.  By contrast mission-awareness involves goal-directed behavior on a specific task or related sets of tasks.

	\subsection{Situational-awareness}
	
	\subsubsection{Main aspects of situational-awareness}
	Embodied perception, that is, situational-awareness by embodied agents, involves two main aspects: the awareness of the environment and awareness of other strategic agents.

	\textbf{(a) Awareness of the environment and the general context}

	The most important task in embodied AI research is aimed at enhancing the situational-awareness of agents – their ability to make sense of the real world in a way that allows them to interact with it and carry out actions towards the achievement of specific goals. Perception in the context of embodied cognition entails not only an understanding of the current state of the world and the processes occurring in it, but also an understanding of how the environment will change in the near and distant future as a result of various factors, including, most importantly, the effects of the actions of the agent and/or other agents. For humans, situational-awareness is a result of knowledge acquired through learning and experience, instincts and innate knowledge transmitted through genes, as well as “on the fly” information provided by other humans and intelligent systems.  Intelligent agents based on LLMs also possess similar attributes. For example, the core model itself is a knowledge base for commonsense generic knowledge about the world \cite{singhal2023large,liu2024gpt,chang2024survey,tu2024towards}. In addition, specific information about the world can be acquired through various means (see discussions in subsection 2.2)

	\textbf{(b) Awareness of users and other agents}
	
	Most real-world settings are complex multi-agent environments, where the behavior of agents is influenced not only by static and unintelligent inanimate objects and variables, but also by the intelligent and purposeful actions of other agents which can be cooperative or competitive at a given time.  In such an environment, behavioral outcomes depend on the goals, intelligence and the overall competences of other agents. However, it is often not possible to directly observe the properties – i.e., access the goals and strategies – of other agents. These properties are inferred from the actions and reactions of the agents in the course of interaction. Prior knowledge about their behavior can also be incorporated in the LLM model. Specialized datasets and finetuning methods can also endow LLM models with knowledge about the behavior or other agents.  Some works incorporate specialized cognitive modules to infer various attributes about other agents, including their believes, intentions, knowledge level and general state of mind \cite{cross2024hypothetical,chernyavskiy2024applying}. Hypothetical Minds (HM) \cite{cross2024hypothetical} observes the action history of other agents and leverage the information to predict their strategies and output a high-level description in natural language which can then be utilized to refine the model’s (HM’s) own behavior.
	

	\begin{figure*}[!htb]
		\vspace {-1mm}
		\centering
		\includegraphics[width=0.90 \linewidth]{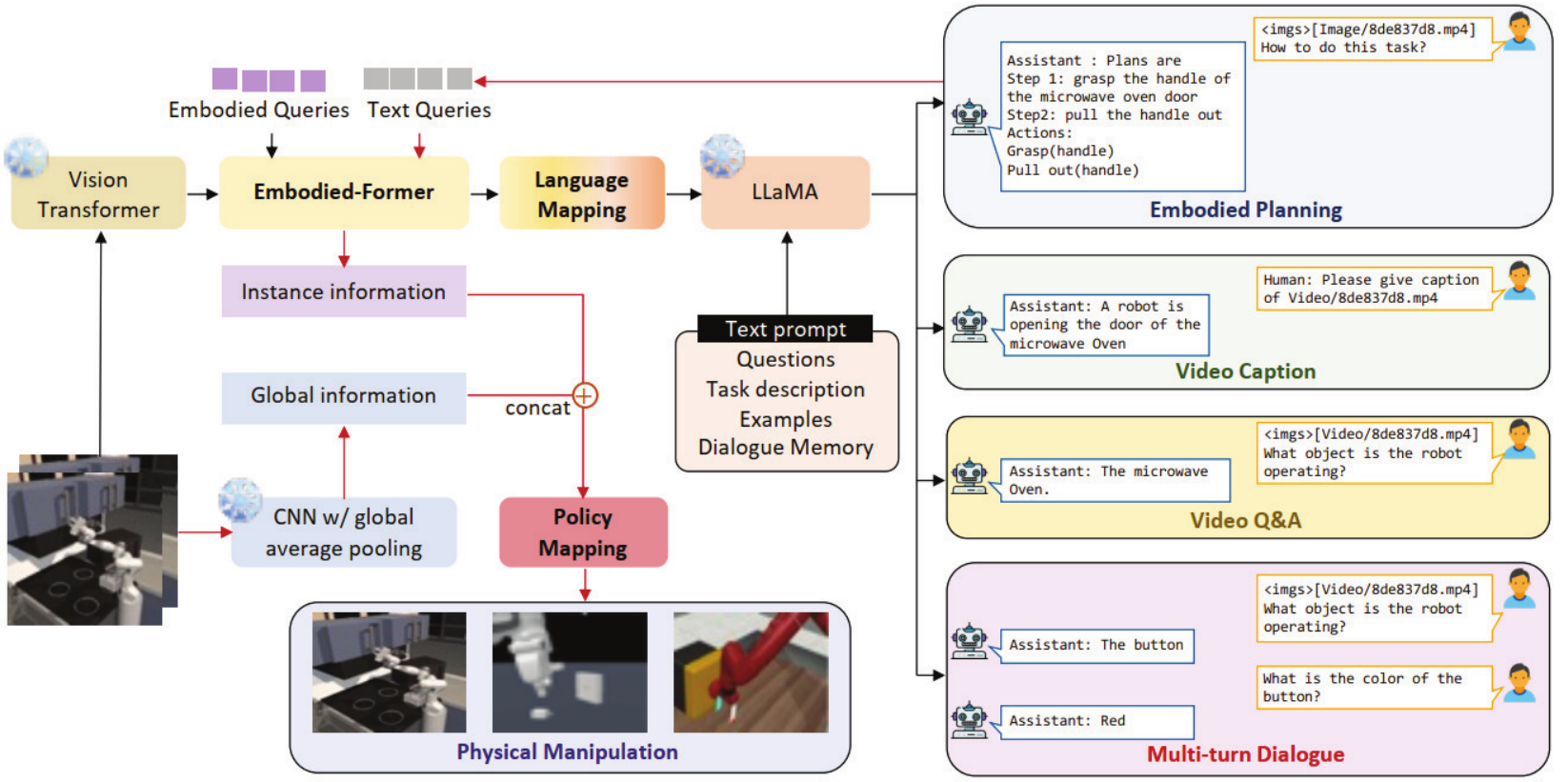} \vspace {-3mm}
		\caption{A simplified representation of EmbodiedGPT \cite{mu2024embodiedgpt}. The framework utilizes a large-scale egocentric, EgoCOT—curated as part of the work, to teach agents a wide range of embodied skills, including video captioning, visual question answering, multi-turn dialog as well as navigation and object manipulation in the physical world. It consists of four integrated components: (a) a vision-transformer to encode visual information from observations; (b) a custom submodule, so named Embodied-Former, to map input text and images (i.e., embodied instructions and visual information), and to generate relevant features for embodied, high-level planning and low-level control tasks; (c) a large language model to perform language-related tasks (e.g., image captioning, planning and embodied question answering); (d) a so-called policy network that generates low-level actions from the features learned by the Embodied-Former submodule. These actions allow the agent to physically interact with the real world using its actuators. Chain of thought approach is used to generate task-relevant goals from prompts.   
		}\label{fig:1E_EmbodiedGPT}
	\end{figure*}
	

	\subsubsection{Approaches to realizing situational-awareness in embodied LLMs}
	
	\textbf{(a) Physical agents in real world environments}
	
	The most straightforward approach to realize embodiment in LLMs is to design and implement embodied agents in the form of robots with appropriate sensing modalities and then integrate the advanced language understanding capabilities with the robot's physical and sensory mechanisms. Embodied generalist agents must perform multiple tasks at the same time: perception, planning, navigation, object manipulation, natural language communication, physical interaction with humans and other AI agents, as well as and low-level control tasks. In principle, generalist embodied agents can be trained on specially-curated embodied datasets such as EgoExoLearn \cite{huang2024egoexolearn}, Holoassist \cite{wang2023holoassist}, EgoTracks \cite{tang2024egotracks} and EgoChoir \cite{yang2024egochoir} in an end-to-end manner. Special embodied multimodal models such as EmbodiedGPT \cite{mu2024embodiedgpt} (see Figure \ref{fig:1E_EmbodiedGPT}), PaLM-E \cite{driess2023palm} and AlanaVLM \cite{suglia2024alanavlm} are trained on these types of multisensory embodied dataset. The data commonly consist of egocentric datasets containing videos of humans performing diverse actions in different settings. 
	
	The actions are typically aligned with context-relevant language descriptions. In addition, they sometimes often audio and other sensory information. In order to ensure that the embodied datasets are as realistic and informative as possible, some works (e.g., \cite{xu2023towards,fan2024emhi,ashok2024egocentric+}) leverage wearable sensors like accelerometers, inertial measurement units (IMUs), global navigation satellite systems (GNSS), head-mounted displays (HMDs) and gyroscopes to capture additional information (e.g., location, orientation, pose, etc.) about the environments, objects, humans and the activities. Thus, the task of training a multimodal model is to learn a common representation for these multiple sensory information types. While this approach has demonstrated impressive capabilities for robots and embodied autonomous agents, it is exorbitantly costly and time-consuming to collect such datasets. 
	
	Since it is often difficult and costly to develop and train language models from scratch for general-purpose multisensory embodied robotic systems, most works typically fine-tune pretrained multimodal large language models with task-specific datasets. That is, the realization of physical systems for embodied AI involves adapting the model to handle specific tasks and interactions – e.g., navigation \cite{wang2022towards,bigazzi2021out,gao2024vision}, manipulation \cite{li2024manipllm,zhang2024empowering}, human-machine dialogue \cite{kalinowska2023embodied,shidara2024adapting} – that are relevant to the physical capabilities and sensory inputs of the target embodied AI system. For instance, Palm-E \cite{driess2023palm} is specifically designed for kitchen settings. Therefore, the most common tasks it performs are navigation in the kitchen environment, recognizing household objects, picking and placing cooking utensils and other objects, assisting with general chores relating to cooking, cleaning, and serving food. Thus, the approaches are usually domain-specific addressing a restricted set of situations.  
	Owing to the difficulty in curating sufficiently large and diverse real data for embodied tasks, many works \cite{taran2024benchmarking,leonardi2022egocentric,birlo2024hup} train large multimodal language models using synthetic datasets or augment real datasets with synthetically-generated egocentric data. Dedicated frameworks for generating (e.g., LEAP \cite{dessalene2023leap}, EgoGen \cite{li2024egogen}) or annotating (e.g., PARSE-Ego4D \cite{abreu2024parse}) synthetic egocentric data have been proposed. Generally, the target tasks and specific interactions the embodied AI needs to handle (e.g., navigation, manipulation, human-machine dialogue) are predetermined and a suitable dataset is selected or generated. While this workaround effectively mitigates the data curation challenge for specific embodied tasks, it is still difficult to extend these models to general, open-ended, long-horizon tasks. This is largely because current synthetic datasets, like their real counterparts, capture short, independent video snippets containing only partial and local information about the underlying
	Physical environment and tasks. It is particularly challenging to handle multi-agent systems \cite{wang2024survey,zhang2024proagent,guo2024large} in complex environments where multiple factors interact over long-horizons. To address this limitation, some recent approaches propose to combine multiple specialized embodied modules to perform specific tasks \cite{zhang2023building,hassouna2024llm}.

	However, in terms of achieving AGI, this approach is still extremely limited. Firstly, the datasets are often static, with less opportunities for learning rich representations and complex skills. As these datasets are not interactive, agents can only make passive observations and process these observations as sensory signals for acting on the world. Secondly, agents cannot “live” in these environments and have experiences from first-person perspective. Moreover, training with such static datasets is fundamentally different from learning in the real world, where agents’ observations are a result of their own, mostly deliberate, actions – i.e., the agent controls the data it receives through interactions with the environment. A promising workaround is to train the agent model in a virtual world, a more complete simulation environment, and then transfer to the real world.

	\textbf{(b) Simulated agents and virtual environments for embodied AI systems}
	
	A promising approach to simplifying the difficulty of developing and training embodied agents in the real world, is to create and train virtual agents in simulated 3D digital environments. This provides a low-risk, quick and cheap means to learn about the world. In a simulated environment, agents can also learn from humans through human-computer interaction (HCI) interfaces \cite{pan2024human,zhang2023large,padmanabha2024voicepilot}. They can also learn from the experiences of other agents through observation or interaction with them \cite{sun2024llm,hu2023enabling,liu2023dynamic}. This is consistent with the way human learn in the real world. This shared observations and knowledge exponentially enhances the capabilities of the intelligent agents. 
	
	The trained model learned in the virtual environment can then be transferred and finetuned for agents in the real world. Using this approach, sophisticated embodied agents can be effectively trained for complex, dynamic and unknown environments without requiring carefully-curated datasets or prior knowledge about the structure, sensory modalities and functions of the agent itself. It should be noted that autonomous agents such as chatbots, avatars, virtual medical assistants and conversational recommender agents can fully function as virtual agents instead deploying in cyberphysical systems. Such agents operating in virtual mode can still be considered as embodied in the sense that in their context (i.e., in the virtual sense) they can be endowed with most of the attributes of embodied intelligence, including virtual bodies, internal model of behavior, sensing and actuation capabilities that allow them to interact with the physical environment and receive feedback about the interactions. 
	
	To facilitate general intelligence, the virtual environment must meet the following vital requirements:
	
	\begin{itemize}

		\item Large scale, with the possibility of extension
		
		\item Computationally efficient
		
		\item Rich and informative
		
		\item Adequate diversity and variability
		
		\item Realistic and physically-plausible
	\end{itemize}

	\begin{figure}[H]
		\vspace {-3mm}
		\centering
		\includegraphics[width=1.0 \linewidth]{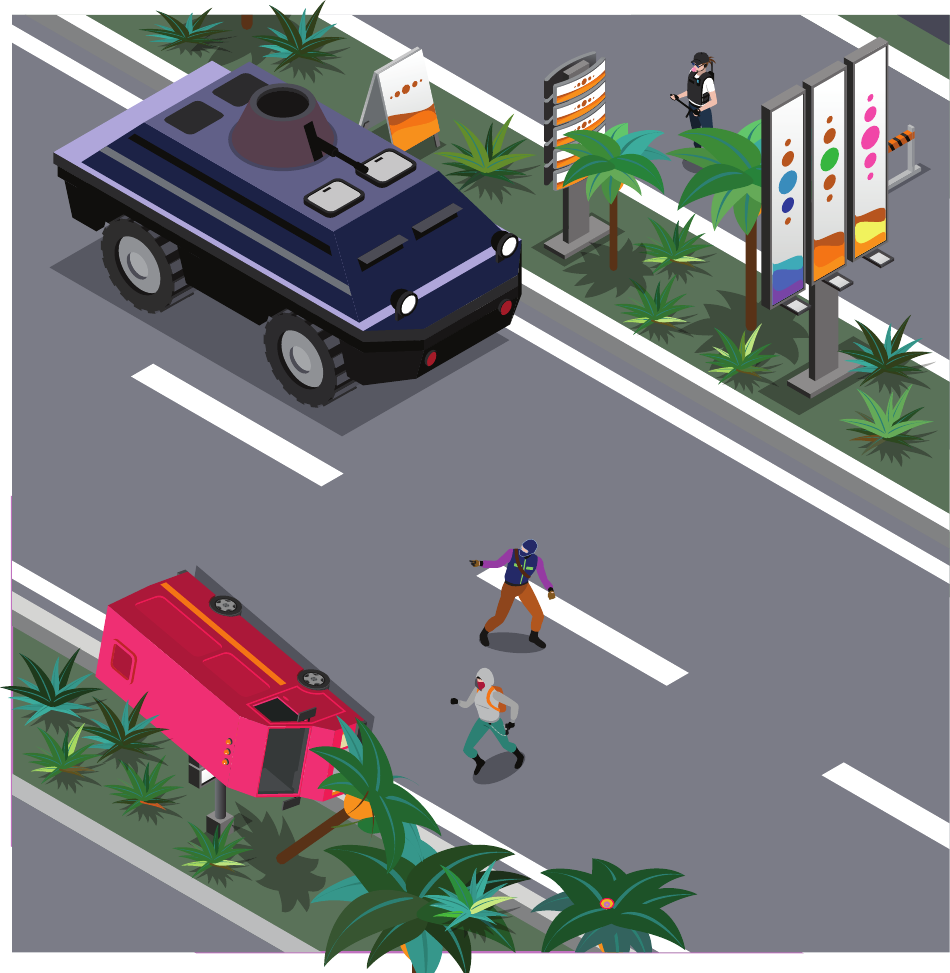} \vspace {-3mm}
		\caption{A practical case where situatedness (situational-awareness coupled with self-awareness) is needed: the red mini bus has to swerve the pedestrians to avoid knocking them down. Situatedness helps AI-based systems such as autonomous vehicles to adhere to acceptable behavior in society and also enables them to avoid serious incidents while incurring minimal penalty (e.g.,injury to occupants, damage of vehicle). In this particular case, situational-awareness allows the AI agent to understand the scene and know where to move to so as to avoid running into another danger while swerving the pedestrians. On the other hand, self-awareness allows the AI-driven vehicle to consider its own physical constraints in order to perform safe maneuvers.
		}\label{fig:4E_Awareness_Situational}
	\end{figure}


	\begin{figure*}[!htb]
		\vspace {-1mm}
		\centering
		\includegraphics[width=0.90 \linewidth]{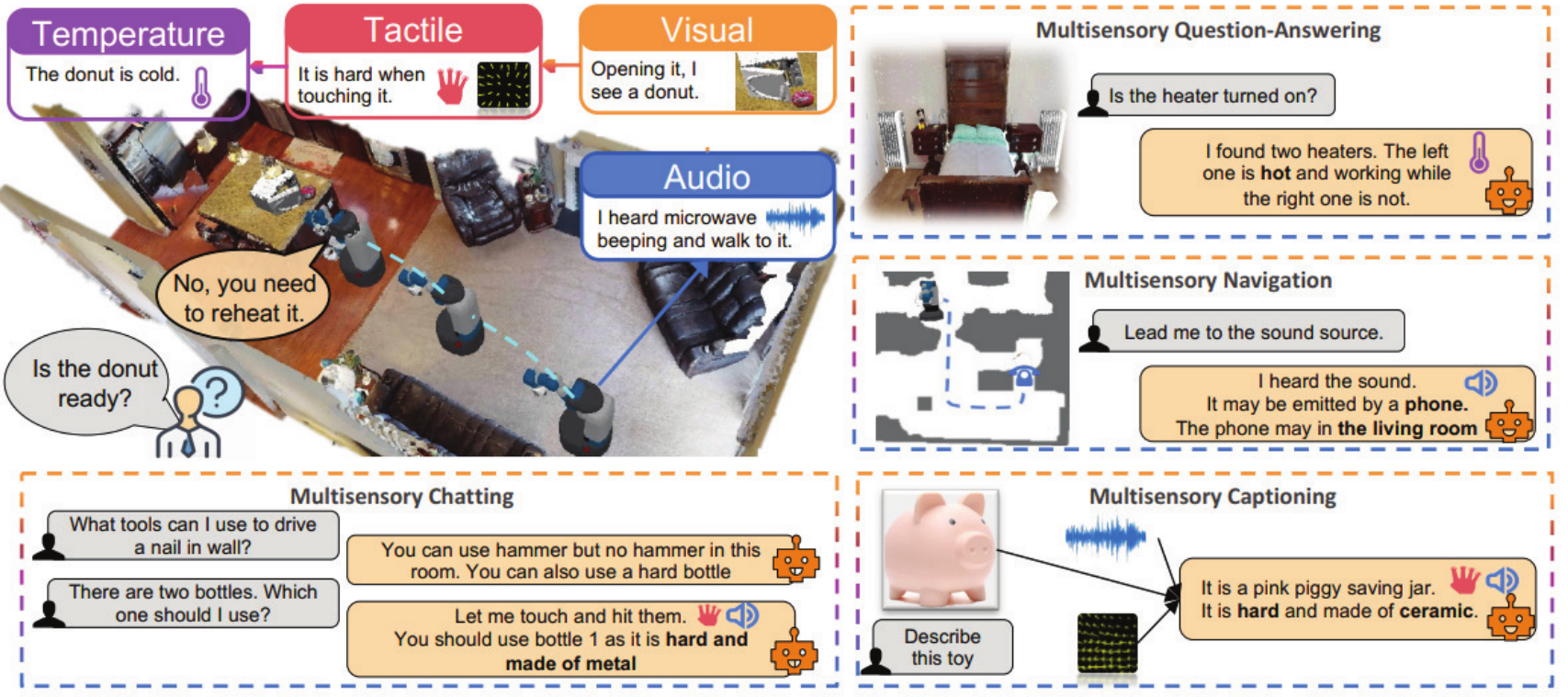} \vspace {-3mm}
		\caption{MultiPLY \cite{hong2024multiply}, a state-of-the-art embodied LLM trained on simulated worlds, supports a vast array of sensory modalities, including textual, visual, thermal, audio and tactile. By jointly encoding context information from multiple sensory modalities and learning the relationships among these diverse representations, the framework achieves advanced capabilities on multiple open-domain tasks such as task planning, tool use, multimodal dialogue, video captioning, question answering, spatial reasoning and navigation.  
		}\label{fig:2E_MultiPLY}
	\end{figure*}
	
	
	\textbf{Common types of simulated worlds for training embodied agents}
	
	Simulated agents and virtual environments for embodied AI can be created in different ways. Some of the common approaches are based on using 1) 3D game engines and 3D graphics tools— e.g., \cite{zhao2025see,gong2023mindagent}; 2) realistic physics simulators— e.g., \cite{coumans2016pybullet,yu2024mhrc,bhat2024grounding}; 3) extended reality (XR) technologies—e.g., \cite{buldu2024cuify,de2024llmr}; and 4) generative AI techniques such as LLMs and VLMs—e.g., \cite{zala2024envgen,wang2024can}. We briefly describe each of these methods in the following paragraphs.

	\textit{(i) Game engines and 3D graphics}
	
	One of the most popular methods for developing simulated embodied agents in virtual environments is by the use of game engines 3D graphics tools. These tools can simulate realistic environments with dynamic conditions (e.g., rainy, bright, dark and foggy weather). These environments support virtual sensors and interactive objects that allow the agents to learn useful skills, affordances and associated constraints that mimic their real-world versions. 
	
	For a number of reasons, there is a big incentive to use tools such as such 3D game engines. Firstly, it easy to create large-scale, realistic environments with these tools. Secondly, already-made generic environments that can be used for training LLM agents are widely available. Popular 3D environmnts like AirSim \cite{shah2018airsim}, AI2-THOR \cite{kolve2017ai2} and Carla \cite{dosovitskiy2017carla} have been created with the Unreal Engine. In turn, tools for training LLM agents can be derived from these 3D simulation models. For instance, LLM-based multi-agent environment simulation frameworks such as EAI-SIM \cite{liu2024eai} and AeroVerse \cite{yao2024aeroverse} are based on AirSim \cite{shah2018airsim}. MultiPLY \cite{hong2024multiply} learns by interacting with simulation environments that integrate sensory-coupled 3D virtual objects—that are in turn derived from the Objaverse \cite{deitke2023objaverse} and ObjectFolder \cite{gao2021objectfolder}  datasets—into a large-scale virtual world built around  Habitat-Matterport 3D \cite{ramakrishnan2021habitat}. In addition, non-playable characters (NPCs) \cite{rodrigues2021shriek,mehta2022exploring} created for computer games that (by themselves) exhibit intelligence and interact with the target agents (e.g., see \cite{christiansen2024exploring}) can be imported to existing environments as game assets. They support complex behaviors, long-horizon interactions and can engage in sophisticated storylines.

	One of the main limitations of this approach is the enormous computational requirements for developing world-scale environments. There is often the need to balance realism with game performance, thereby constraining the level of realism that can be achieved. Another major challenge relates to the inadequacy of game engines to effectively model complex, physically-plausible mechanical interactions as the tools are typically optimized for visuals, an important feature in gaming. 
	
	\textit{(ii). Realistic physics simulations}
	To overcome some of the aforementioned shortcomings of game engines in creating realistic simulation environments for training embodied agents, recent works employ physics engines (e.g., PhysicsX, Bullet, Symbody and ODE) \cite{kaup2024review} to create realistic, physically plausible simulators such as PyBullet \cite{coumans2016pybullet}, Isaac Gym\cite{makoviychuk2021isaac}, DIFFTACTILE \cite{si2024difftactile} and for generating virtual agents and environments. For instance, state-of-the-art models for LMM embodied agents such as ROS-LLM \cite{mower2024ros}, LANCAR \cite{shek2023lancar} and MHRC \cite{yu2024mhrc} are based on PyBullet \cite{coumans2016pybullet}. Compared with 3D game engines, this approach provides more controlled, physics-informed environments where agents can interact with objects and phenomena in a way that aligns with real-world behavior. These tools are particularly suited for agent learning approaches based on reinforcement learning paradigm, since the embodied LLM agents can receive realistic rewards or penalties based on their actions in the environment. The realism provided by physics-based simulators helps agents develop more practical, transferable skills that could eventually be applied in real-world scenarios. Complex tasks such as object manipulation, embodied path planning, and interaction with dynamic environments can be learned more effectively with such accurate physics. 
	
	While physics simulation supports more realistic behavior, the approach is inherently expensive. Moreover, it is often not possible to simulate very complex behaviors or phenomena whose underlying mechanisms are unknown. These problems constrain the range of scenarios that can be effectively modeled and the degree of sophistication that can be achieved in any given task.

	\textit{(iii). Simulated virtual worlds in extended reality (XR)}
	Immersive experiences offer more natural settings for AI agents to acquire useful skills as the virtual agents can interact seamlessly with humans and the real world. Agents trained in this type of environment can comprehend complex, multimodal input, including gestures and emotions, and generate contextually appropriate responses \cite{bovo2024embardiment}. Extended reality (XR) tools, especially virtual reality (VR) and mixed reality (MR) techniques, can create immersive, highly interactive 3D environments that accurately simulate real-world behavior. In XR environments, embodied LLM agents can leverage simulated sensorimotor feedback to learn to perceive and act in the world. In mixed reality mode, virtual agents ‘live’ in the real world and can interact seamlessly with the real world as well as with other virtual objects \cite{bovo2024embardiment}.. Such mixed reality agents can directly perceive the real world through sensors and internet of things (IoT) devices \cite{andrade2019extended}. VR worlds can also provide realistic environments with human-looking virtual agents in the form of avatars that interact with and learn from humans. Social XR \cite{wan2024building,rossetti2024social} platforms create settings that allow diverse humans to engage in practical human-centered activities (e.g.,  trading, shopping, etc.) with virtual objects \cite{kurai2024magicitem,hayase2024panotree}. They can simulate human-agent and agent-agent interactions, making it an ideal setting for training agents in social or collaborative tasks.

	While immersive virtual environments have been created using computer graphics tools and game engines, more recently, the use of generative AI techniques have been employed to build entire XR world models \cite{giunchi2024dreamcodevr,izquierdo2024virtual} or to create specific content for existing XR worlds {wang2024systematic}. The power of have LLMs have also be exploited to adapt computer graphics-generated worlds (e.g., in SituationAdapt \cite{li2024situationadapt} and GUI-WORLD \cite{chen2024gui}) to the underlying social setting and physical environmental attributes.


	\begin{figure*}[!htb]
		\vspace {-1mm}
		\centering
		\includegraphics[width=1.0 \linewidth]{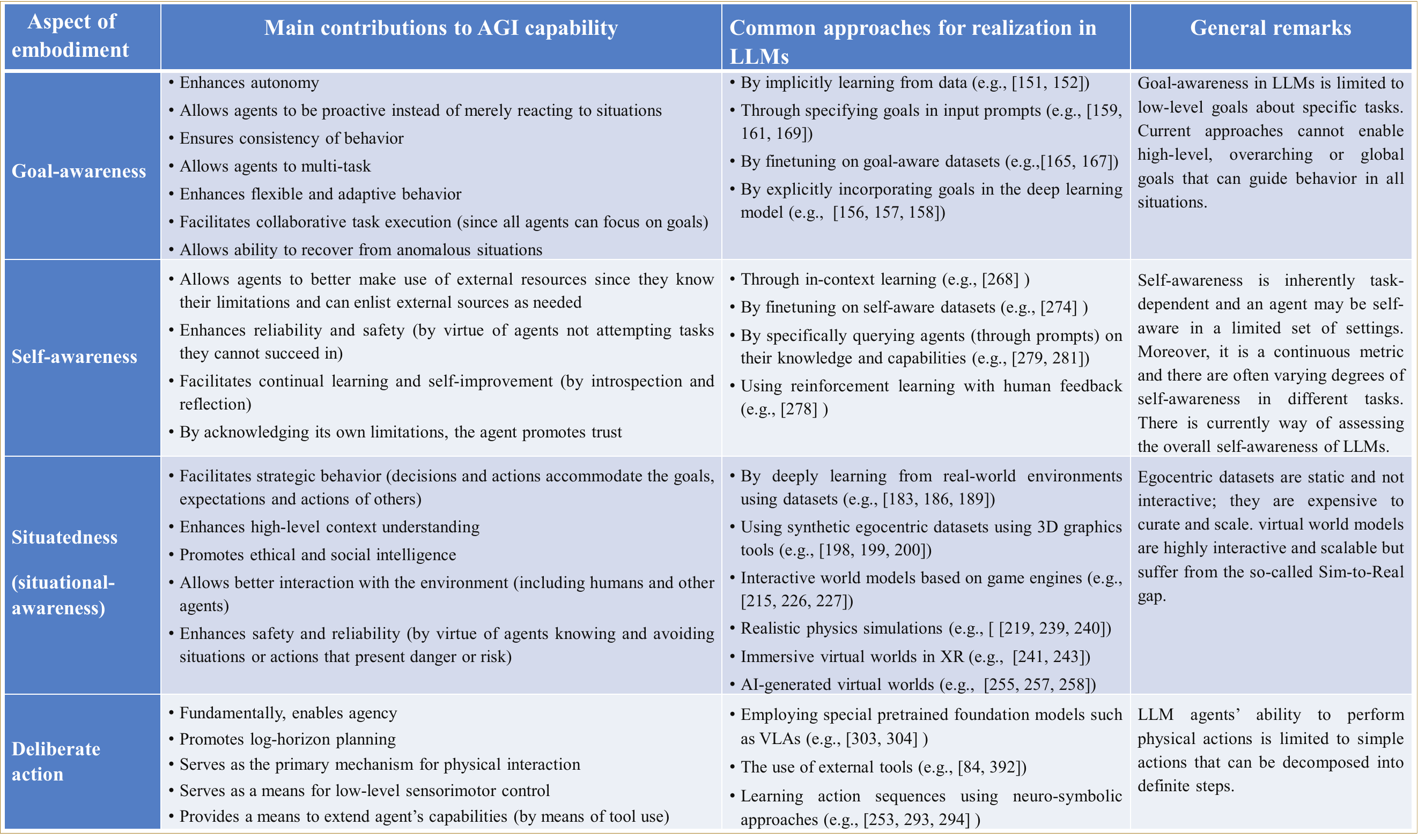} \vspace {-3mm}
		\caption{A summary of the important contributions of the main aspects of embodiment to AGI capabilities and the general approaches to implementing each of these components of embodiment in large language models.   
		}\label{fig:8_All_Summary_Emb}
	\end{figure*}
	

	\textit{(iv) Virtual environments generated by AI (e.g, LLM and VLM)}
	Because of the complexity of simulated virtual environments and the associated high labor cost in their creation, a large number of recent approaches (e.g., \cite{zala2024envgen,wang2024can,huang2023voxposer,wang2023bytesized32,liu2023llm+}) have proposed to circumvent this problem by utilizing pretrained foundation models as world simulators to accurately infer the properties of the world and, hence, produce embodied action plans and predict how different actions alter the world..  With this approach, the LLM frameworks are specially constructed to leverage their rich knowledge to generate embodied training environments that serve as media in which other embodied LLM \cite{yang2024embodied} and VLM \cite{yang2024embodied} agents will be trained. This line of work has been particularly successful in robotics for complex tasks like embodied planning, navigation and manipulation tasks.

	A common approach \cite{hu2024scenecraft,huang2023voxposer,tang2024worldcoder,ma2024llms,huang2023instruct2act} is to produce intermediate code from high-level goals that are specified as natural language instructions which is then used to generate plausible 3D world models that embodied agents interact with. These methods typically utilize the rich prior world knowledge encoded in the LLM to guide code generation. The generated code can further manipulate the world model to produce diverse scenes and environment conditions based on the artificial agents’ desired goals and experiences. Hu et al. \cite{hu2024scenecraft} propose a code generation approach that synthesizes 3D scenes by generating Blender code. To achieve this, they build a scene graph that encodes the geometric relationships and constraints of primitive 3D objects. A specialized VLM module based on GPT4-V, so-named SceneCraft, is then able to leverage the scene graph to generate Python scripts that creates and populate a 3D scene in Blender with relevant objects. SceneMotifCoder \cite{tam2024scenemotifcoder}leverages LLM code generation for open-vocabulary 3D object generation and arrangement in a geometry-aware manner. 
	Tang et al. \cite{tang2024worldcoder} frame the embodiment problem as a model-based reinforcement learning task that leverages prior knowledge in the form of LMM to learn embodied planning and actions with only a few interactions with the environment. The world model in these cases are built from python code with the help of natural language instructions. While this approach seems promising, the practical difficulty of modeling large, complex and dynamic environments in this manner limits the method to relatively simple environments. For instance, for practical realization, Tang et al. \cite{tang2024worldcoder} formulate the virtual scene as a deterministic environment where interactions are episodic. 
	Owing to the difficulties in achieving realistic, physics-based interactions with LLM-generated worlds, further refinements are often employed to ensure physically-plausible behavior \cite{volum2022craft,de2024llmr,numan2024spaceblender,qian2024shape}. For instance, Volum et al. propose a code generation method to synthesize interactive objects and characters for virtual worlds using LLM prompting. Their approach, \textit{Craft an Iron Sword} \cite{volum2022craft}, additionally employs the LLM to infer interaction outcomes and generate plausible response (i.e., in the form of scene manipulation).

	\subsection{Self-awareness}
	Self-awareness is the ability of the AI system to understand its very nature, including its properties, capabilities, limitations, context, and role in interactions with external entities. The body's physical structure influences how biological systems or living organisms process information. Its shape, size, and capabilities constrain and afford certain capabilities and types of actions, which in turn affect the cognitive strategies required to achieve them. The nervous systems of living organisms naturally learn to control the body mechanisms such as muscles and limbs in a body-specific way. This explains, for instance, why humans require extensive training to be able to easily use prosthetic limbs \cite{windrich2016active,farina2023toward}. A self-aware embodied agent can also understand the implications of its actions on other agents (human and artificial) and the environment overall. An AI agent that is both self-and situational-aware is said to be situated. Figure \ref{fig:4E_Awareness_Situational} depicts a typical scenario where situatedness is critical for an AI system to take correct decisions.

	\subsubsection{Self-awareness in generic LLMs }
	Many researchers (e.g., \cite{li2024benchmarking,chen2024persona,wang2024mm}) have investigated self-awareness in LLMs, including the ability know the limit of their own knowledge \cite{yin2023large,wang2023survey,prato2024large}, and to introspectively reflect on decisions and actions \cite{qu2024recursive,liang2024introspective,song2024exploreself},  and adjust behavior. Based on the preliminary evidence, multimodal LLMs are generally considered to possess self-awareness as an emergent ability, an ability that arises spontaneously as a result of the sheer volume of training data. Yin et al. \cite{yin2023large}, for example, show through extensive empirical studies that state-of-the-art LLMs naturally possess a degree of self-awareness about the limit of their own knowledge, i.e., knowing what they do not know. Several other studies confirm this ability \cite{wang2023self,laine2024me,davidson2024self,wen2024perception,pushpanathan2023popular}.

	\subsubsection{Achieving self-awareness in LLMs}
	
	While, out-of-the-box, current state-of-the-art LLMs like GPT-4 still lack true self-awareness in the sense of human cognition, a number of techniques can help to elicit self-awareness. For instance, it has been shown that in-context learning \cite{yin2023large}, reinforcement learning with human in the loop \cite{ouyang2022training} and fine-tuning \cite{laine2024me} can be used to attain some level of self-awareness in LLMs. A common way to achieve self-awareness is to evaluate a model’s outputs for inconsistencies or errors by comparing generated responses with known facts or previous dialogue. In this way, the model can be explicitly prompted about its limitations  \cite{wang2023self,gao2023retrieval}. Instead relying on humans to probe and prompt LLMs about their knowledge or capabilities, a recent approach is to formulate the self-awareness task as an intuitive search question, whereby an embodied agent queries its base LLM on the existing world knowledge about the given situation \cite{cheng2024can,zhu2024bootstrapping,amayuelas2023knowledge}.   In line with this line of work, multiple LLM agents can collaborate to assist one another, through probing or questioning, to reveal  capabilities and inherent weaknesses \cite{feng2024don}. More recent works (e.g., \cite{yao2024seakr,liang2024internal,peng2024self}) have proposed to infer the properties of the LLM models from the hidden representations of their internal states. SEAKR \cite{yao2024seakr} computes a so-called self-aware uncertainty from latent representations of the internal states of the LLM's feedforward network by comparing the consistency score across multiple responses. Self-Controller \cite{peng2024self} incorporates a dedicated submodule, state reflector, that stores state information for assessment. Potentially, these approaches could be extended to address awareness of more pertinent attributes of embodied LLM agents, allowing them to, for example, become cognizant of their physical construction, the mechanisms of action and response, as well as the attendant outcomes of actions and their own physical limitations. A self-aware LLM can recognize when its internal knowledge is insufficient to address a problem, and turn to additional resources, e.g., retrieval-augmented generation \cite{lewis2020retrieval,jiang2023active}. This approach is similar to the way human enlist additional resources to tackle problems they are not capable of handling by themselves. The useful attributes of awareness of both self and the environment in embodied AI systems is better illustrated by Figure \ref{fig:8_All_Summary_Emb}.

	\subsection{Deliberate action}
	Although LLMs are primarily language entities, when endowed with embodiment—e.g., as physical robots, virtual agents, or other interactive systems—they can take deliberate actions in the real world or through virtual or simulated interactions. This capability stems from their ability to comprehend task- or goal-oriented dialogue \cite{mu2024embodiedgpt,wang2024safe,wu2023autogen,liu2022instruction}, formulate step-by-step plans to accomplish tasks or to achieve target goals \cite{xie2023translating,liu2023llm+,valmeekam2023planning} and execute task-appropriate actions according to the predefined plans \cite{graule2024gg}. An embodied agent can discover new affordances and previously unknown properties of objects through deliberate interaction with the environment. This enables it to make decisions or take actions that are more beneficial, empathetic and morally informed.
	Most embodied LLM systems (e.g., \cite{liu2023llm+,silver2024generalized,silver2022pddl}) incorporate dedicated planning and action submodules to handle action execution and interaction with external entities. These specialized modules typically employ representations of action primitives that are related to the agent’s design and capabilities. The action primitives are then encoded in the form of policies (i.e., rules for allowable behavior) \cite{zhai2024fine,raman2022planning,shi2024large} or into action templates \cite{guan2023leveraging,mahdavi2024leveraging} that describe how to respond to various scenarios. Through the ability to use external tools \cite{paranjape2023art}, LLMs can extend their potential to perform various actions.
	 
	Vision-language-action models (VLAs) \cite{gbagbe2024bi,zhen20243d,jang2022bc,lu2024unified,ding2025quar} are a new family of multimodal foundation models specifically designed to execute actions. They jointly learn visual, language and action modalities through end-to-end training. Consequently, they can perceive the environment, interpret instructions, carry out high-level planning and synthesize low-level actions to complete various tasks. VLAs are commonly used in robotic applications. They particularly excel in tasks such as open-world navigation, object manipulation, grasping and interpreting and responding to complex sensorimotor signals, including verbal and nonverbal cues. State-of-the-art Bi-VLA \cite{gbagbe2024bi}, VLAs such as RT-2 \cite{zitkovich2023rt}, Unified-IO 2 \cite{lu2024unified}, QUAR-VLA \cite{ding2025quar} and 3D-VLA \cite{zhen20243d} can perform a wide range of complex activities in open-domain settings.

	\section{Symbol grounding}
	
	\subsection{Basic idea of symbol grounding}
\begin{figure}[H]
		\vspace {-3mm}
		\centering
		\includegraphics[width=1.0 \linewidth]{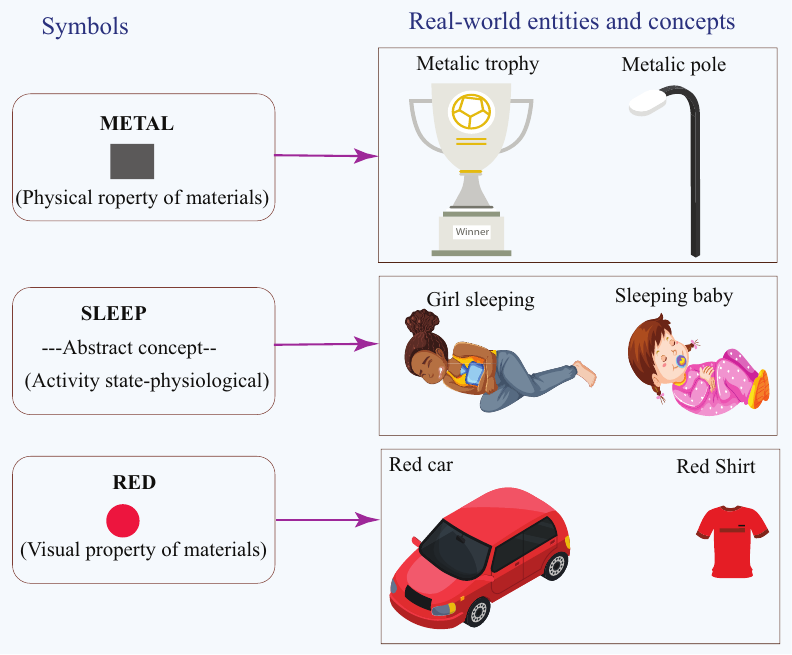} \vspace {-3mm}
		\caption{Human cognition relies on associating abstract mental representations, or symbols (e.g., words), with entities, concepts and phenomena in the real world. The symbol grounding system allows the internal cognitive system to access the external world. In this manner, internal representations acquire meanings that are invariant in a given sense, and can therefore identify referents (i.e., the objects or categories referred to) in different contexts. In the same way, grounded artificial cognition aims to connect abstract computational representations to actual objects and concepts with respect to some concrete interpretation. 
		}\label{fig:9_Grounding_Descrete}
	\end{figure}

\begin{figure}[H]
		\vspace {-3mm}
		\centering
		\includegraphics[width=1.0 \linewidth]{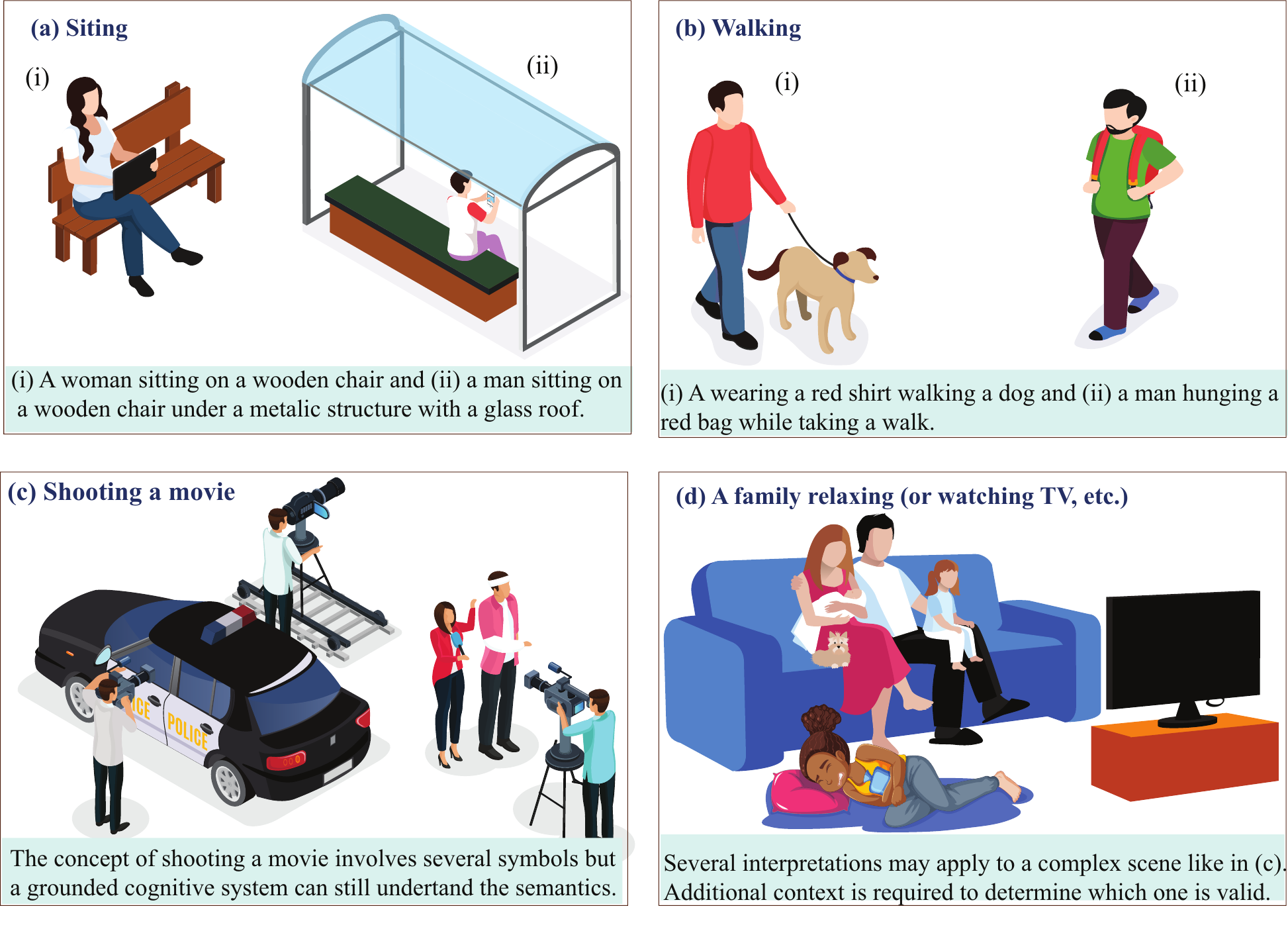} \vspace {-3mm}
		\caption{The grounding mechanism allows intelligent systems to represent and manipulate cognitive information in a hierarchical manner by combining different symbols to form more complex, composite representations. In (a), for example, the symbols "metal", "wood", "chair", "structure" all provide context to describe the high-level concept "sit". The semantic content in (b) is the activity "to walk". (c) and (d) are more complex scenes consisting of several hierarchies of symbols but they still represent very simple semantic contents as the labels indicate. 
		}\label{fig:10_Grounding_Composite}
	\end{figure}

	Symbol grounding, or simply grounding, relates to the ability of AI systems to connect the abstract internal representations of concepts in computational models to their real-world equivalents. In its basic form, the grounding problem essentially involves specifying a set of primitive symbols, defining their semantic connotation and postulating rules for manipulating them. The rules that govern symbol manipulation are purely syntactic in nature, and are independent on the assigned meaning (i.e., the real-world, physical interpretation) of the symbols \cite{harnad1990symbol,vogt2002physical}. The symbols themselves are abstract primitive entities that are treated as atomic tokens that can be combined into composite tokens to encode higher-level concepts \cite{taddeo2005solving}. The symbol system is supposed to be semantically interpretable at all levels of representation \cite{harnad1990symbol} (see illustrations and further explanation in Figures \ref{fig:9_Grounding_Descrete} and \ref{fig:10_Grounding_Composite}). Symbol systems are, thus, patterns of information that provide access to the external world. Newell and Simon \cite{newell2007computer} hypothesized that physical symbol systems are not only necessary but also sufficient for intelligence.

	Methods of grounding in artificial intelligence and large language models have been inspired by the way the human brain processes and associates sensorimotor information to the external world \cite{cangelosi2005approaches,macdorman1999grounding,pecher2011abstract, pylyshyn1980computation}. Psychologists have long argued that the human mind itself relies on a symbolic system of representation and manipulation of information in mental processes (see \cite{cangelosi2005approaches,pylyshyn1980computation,searle1980minds,fodor1983modularity,barsalou2008grounded}). Per this view, cognitive phenomena that influence human perception and behavior, including vision, language, emotions, thoughts, perspectives and beliefs, are governed by symbol processing \cite{fodor1980methodological,meins2013mind}. It is worth noting that a large class of symbols do not relate to physical properties of the world but rather to abstract concepts. For example, symbols such as “happy”, “innovation’, “clever” and fascination” are merely concepts that describe high-level phenomena. However, humans are still able to effortlessly connect these symbols with their appropriate semantic contexts. When presented with images of people or even animals, for instance, humans can correctly classify them by their emotional state.
	
	\subsection{Grounding as a bridge between the digital world and reality}
	Language uses symbols (digits, words, lexical concepts, etc.) represent humans’ understanding of various ideas about objects and concepts in the world: their essence, properties, relations, and possible actions that can be performed on them by an agent. The goal is to provide more meaningful and rich contexts of the real world to facilitate a better understanding of and allow interaction with the external environment by establishing the correct relationship between the abstract symbols captured internally in AI models, and the physical world they seek to represent. In essence, grounding aims to bridge the inherent semantic gap that exists between artificial intelligence and the real world. This allows AI systems to “make sense of” inputs from the environment, thereby enhancing their situational awareness and task-appropriate behavior \cite{bajaj2024grounding}. 
	
	\subsection{General approaches to symbol grounding in AI}
	
	Classical techniques for symbolic grounding utilize explicit representations, with fixed rules and ontologies to describe the relationships and properties of the abstract concepts and physical entities involved. For instance, mathematical operations based on variable binding techniques \cite{smolensky1990tensor,frady2021variable} and logic rules \cite{glanois2022neuro,sen2021combining,belle2020symbolic}are often used for symbol manipulation.
	The key advantage of this class of methods is the (increased) transparency and interpretability of the resulting models. However, the approach is highly restrictive as it requires all situations to be anticipated in advance and handled appropriately.  Furthermore, it is challenging to (move) from fixed, structured representation of symbols to high-level cognitive tasks like perception and reasoning in the open world. Another layer of difficulty lies in the ability to unambiguously and reliably ground fuzzy concepts related to human social relationships and interactions as these are often have strong in cultural contexts and lack consistent interpretations. In these scenarios symbolic manipulation techniques are generally incapable of adequately processing cognitive information as high-level rules often fail to capture contextual nuances and the symbols themselves tend to have differing interpretations, leading to unpredictable or inconsistent inference. Because of the serious limitations of the analytical techniques based on fixed symbols and logical rules \cite{correia2014logical}, probabilistic graphical models \cite{tellex2011approaching,kollar2013generalized} and knowledge graphs \cite{guo2016jointly,chaudhuri2021grounding,ji2021survey}  have become more viable alternatives as a result of their flexibility, better representation power and scalability. These recent methods, the so-called neuro-symbolic techniques \cite{de2011neural,liu2024beyond,cheng2024neural,demeter2020just}, employ primitive entities as representation priors but utilize artificial neural networks to learn the relationships and properties of the symbols. This approach has proven effective but also suffers from poor scalability. Another recent approach, neuro-symbolic grounding \cite{li2024softened,zellers2021piglet,hsu2023ns3d, mao2019neuro }, seeks to ground primitive symbols implicitly by learning the semantic connections of abstract symbols and the real world with the help of neural networks. Methods for learning symbol representations implicitly end-to-end from data without relying on explicit primitives have also been proposed \cite{deng2023transvg++,pavlick2023symbols,huang2024grounded,li2024groundinggpt}.

	\subsection{Approaches to grounding in LLMs}
	We discuss the main approaches for symbol grounding in large language models in the remaining subsections. A detailed summary of these approaches is presented in Table \ref{tab:T1_Grounding_approaches}.
	
	\subsubsection{Grounding LLMs with knowledge graphs}
	In large language models, a common way to capture expressive relationships between various entities, in this case, between abstract symbols and real-world entities, is by the use of knowledge graphs (KGs) \cite{chen2020review,fensel2020introduction,wang2017knowledge}. Knowledge graphs represent words as nodes in a visual tree-like structure known as a graph or semantic network. These words represent individual objects, object categories, events and concepts. The relationships that exist among the various words are described by edges connecting the nodes.  Through this mechanism KGs can store a large volume of explicit knowledge that is grounded in real world. For this reason, they have been proposed to mitigate common problems such as hallucination and provide a means to internalize physically-grounded knowledge in LLMs \cite{rosset2020knowledge,liu2020k,sun2019ernie,pan2024unifying}. This reduces the need for very large training data and, hence, saves time and reduces training cost. Besides, in contrast to pure neural architectures, the structured knowledge in KGs encodes explicit relationships and is therefore more semantically meaningful and suitable for emergent tasks such as reasoning and planning.

	While LLMs augmented with KGs can enhance the inferential capabilities of the LLMs, manually building the KGs is a nontrivial task \cite{zhou2020survey}. Consequently, with their extensive world knowledge, LLMs have, in turn, also been proposed to build or enrich KGs (see \cite{zhang2024extract,zhang2020pretrain, abu2024knowledge,pan2023large}). Thus, these two classes of methods, LLMs and KGs, can be integrated in a way that allows them to mutually enhance each another. This capability presents a promising prospect  for symbol grounding as KGs are incorporated in the LLM framework to improve its performance, while at the same time the resulting LMM helps to extend and refine the KG with additional knowledge to even produce better outputs. This could in turn generate even better content for the graph, and so on. Some recent works \cite{xu2024llm,wang2021kepler,khorashadizadeh2024research}
	are already exploring this approach. 
	
	\begin{table*}[h!]
		
			\caption{A summary of the main approaches for grounding large language models}
			\label{tab:T1_Grounding_approaches}
			
			\scalebox{0.75}{
			\begin{tabular}{@{}lllll@{}}
				\toprule
				\textbf{General approach}                                           & \textbf{Description}                                                                                                                                                                                                                                                                                        & \textbf{Rep. works}                              & \textbf{Main strengths}                                                                                                                                                                                       & \textbf{Weaknesses}                                                                                                                                                                    \\ \midrule
				Knowledge graphs                                                    & \begin{tabular}[c]{@{}l@{}}Represents the relationships between\\  symbols  and the actual entities they\\  represent in a   structured form.\end{tabular}                                                                                                                                                  & \cite{rosset2020knowledge,sun2019ernie} & \begin{tabular}[c]{@{}l@{}}Can naturally handle hierarchical relationships; \\ easy to integrate into LLMs; highly transparent.\end{tabular}                                                                  & \begin{tabular}[c]{@{}l@{}}Laborious process; difficult to represent \\ ambiguous or fuzzy concepts.\end{tabular}                                                                      \\ \\
				\begin{tabular}[c]{@{}l@{}}Ontology-driven\\ prompting\end{tabular} & \begin{tabular}[c]{@{}l@{}}Utilizes in-context learning (high-level\\ instructions as input prompts) to explic-\\ itly ground symbols.\end{tabular}                                                                                                                                                         & \cite{ccoplu2024prompt,palagin2023ontochatgpt}   & \begin{tabular}[c]{@{}l@{}}Can be used to refine already-grounded symbols;\\ can be used with any of the other methods; reliable \\ since grounding is explicit.\end{tabular}                                 & \begin{tabular}[c]{@{}l@{}}Cannot accomplish grounding exhaustively;\\ not  scalable; requires knowledge about the \\ underlying   concepts and relationships.\end{tabular}            \\ \\
				\begin{tabular}[c]{@{}l@{}}Vector space \\ embeddings\end{tabular}  & \begin{tabular}[c]{@{}l@{}}Deeply-learns and encodes relationships\\ in the feature space.\end{tabular}                                                                                                                                                                                                     & \cite{zhang2024word,liu2024lang2ltl}             & \begin{tabular}[c]{@{}l@{}}Highly scalable; can learn relationships that aren’t \\ known by the human developers; very simple and\\ easy to implement.\end{tabular}                                           & \begin{tabular}[c]{@{}l@{}}Requires large and high-quality data; may learn\\ spurious relationships;  opaque and difficult to \\ diagnose.\end{tabular}                                \\ \\
				Active exploration                                                  & \begin{tabular}[c]{@{}l@{}}Methods whereby symbols are grounded\\ by embodiment mechanism through inter-\\ action with the world and experiencing the\\ behavior of objects and phenomena and the\\  effects of  actions.   Mostly relies on RL \\ approaches to learn useful representations.\end{tabular} & \cite{hsu2024grounding,carta2023grounding}       & \begin{tabular}[c]{@{}l@{}}There is direct coupling of cognitive information\\ with the real world;  learned representations  are\\  physically-plausible.\end{tabular}                                       & \begin{tabular}[c]{@{}l@{}}Expensive and time-consuming to implement; may\\  produce incomplete connections when used alone;\\ cannot work   on purely abstract concepts.\end{tabular} \\  \\
				Generative AI                                                       & \begin{tabular}[c]{@{}l@{}}Leverages generative AI models ( LLM,\\ GAN, VAE, VLA ,VLM) are employed\\ to  synthesize the patterns and   relation-\\ ships of symbols and referents.\end{tabular}                                                                                                            & \cite{zhang2020pretrain,abu2024knowledge}        & \begin{tabular}[c]{@{}l@{}}Does not require prior knowledge about relationships;\\  extremely scalable; very easy to  implement.\end{tabular}                                                                 & \begin{tabular}[c]{@{}l@{}}Not transparent; prone to fake connections (e.g., \\ hallucinations); not easy to verify; representations\\ may be unreliable.\end{tabular}               \\  \\
				External knowledge                                                  & \begin{tabular}[c]{@{}l@{}}Utilizes readily-available  knowledge\\ about relationships contained in external\\ knowledge bases (e.g., through RAG) to\\  ground symbols in LLMs.\end{tabular}                                                                                                               & \cite{lewis2020retrieval,zhao2024retrieval}      & \begin{tabular}[c]{@{}l@{}}Leverages a wide variety of existing knowledge;\\ can augment other approaches; provides practically\\ unlimited scope of domains and tasks for grounding \\ symbols.\end{tabular} & \begin{tabular}[c]{@{}l@{}}May introduce inconsistences as a result of different\\ representation schemes; external information may be\\ subject to malicious attacks.\end{tabular}    \\ \bottomrule
			\end{tabular}
		} 
	\end{table*}
	
	\subsubsection{Grounding LLMs by ontology-driven prompting }
	Prompting techniques have been used to steer LLMs to generate more nuanced, contextually appropriate responses. The technique utilizes user-supplied instructions or specific examples (i.e., input-output pairs) at the inference stage. The process does not affect the learned model parameters and also avoids costly re-training or finetuning procedures. This form of adaptation, known as in-context learning, can effectively ground and align model inferences with real-world context according to user needs.  More recently, instead of directly inputting human-readable instructions as prompts, a large number of works (e.g.,
	\cite{ccoplu2024prompt,lippolis2024ontogenia,palagin2023ontochatgpt,wang2024llm,giglou2024llms4om}) have sought to leverage ontologies as symbolically grounded knowledgebases that provide context-relevant prompts in an automated way to guide the model on how to effectively deal with specific situations. The ontology engine is created by a formal specification of facts, rules as well as entities, categories, properties, and relations between them. As an alternative to building the symbolic system (i.e., ontology) manually, some works (e.g.,  \cite{babaei2023llms4ol,da2024toward} have proposed to exploit LMMs to create or enhance ontologies. Different types of operations to generate new knowledge from established facts and rules in specific contexts.

	\subsubsection{End-to-end grounding through embedding}
	
	In LLMs, the symbol grounding problem can be solved by implicitly modeling the meanings of and the associations between learned concepts in the high dimensional vector space \cite{camacho2018word,tennenholtz2023demystifying,lyre2024understanding}. In the vector space, symbols such as words and visual concepts are encoded based on the contexts in which they frequently occur and how they relate to other symbols. Researchers have devised techniques that exploit this representation to associate learned embeddings to the actual objects, perceptual experiences, actions, or concepts in the real world \cite{cohen2024survey,zhang2024word,liu2024lang2ltl,vig2020investigating}. These embeddings can also establish the semantic relationships with other concepts. For example, the phrase “Toyota Landcruiser” can be connected to “Car”, “Vehicle”, “Transportation”, etc. Moreover, symbol manipulation procedures can also be learned in an end-to-end manner using the expressive power of the underlying neural networks.
	
	While this class of techniques is more scalable and relatively simple to implement compared with the other grounding methods covered here, it is important to note that the approaches may not yield precise mappings in some cases. It is also not often possible to determine which symbols are not correctly grounded. For these reasons, grounding methods that leverage vector embeddings typically suffer from issues of trustworthiness and lack of explainability.

	\begin{figure}[H]
		\vspace {-3mm}
		\centering
		\includegraphics[width=1.0 \linewidth]{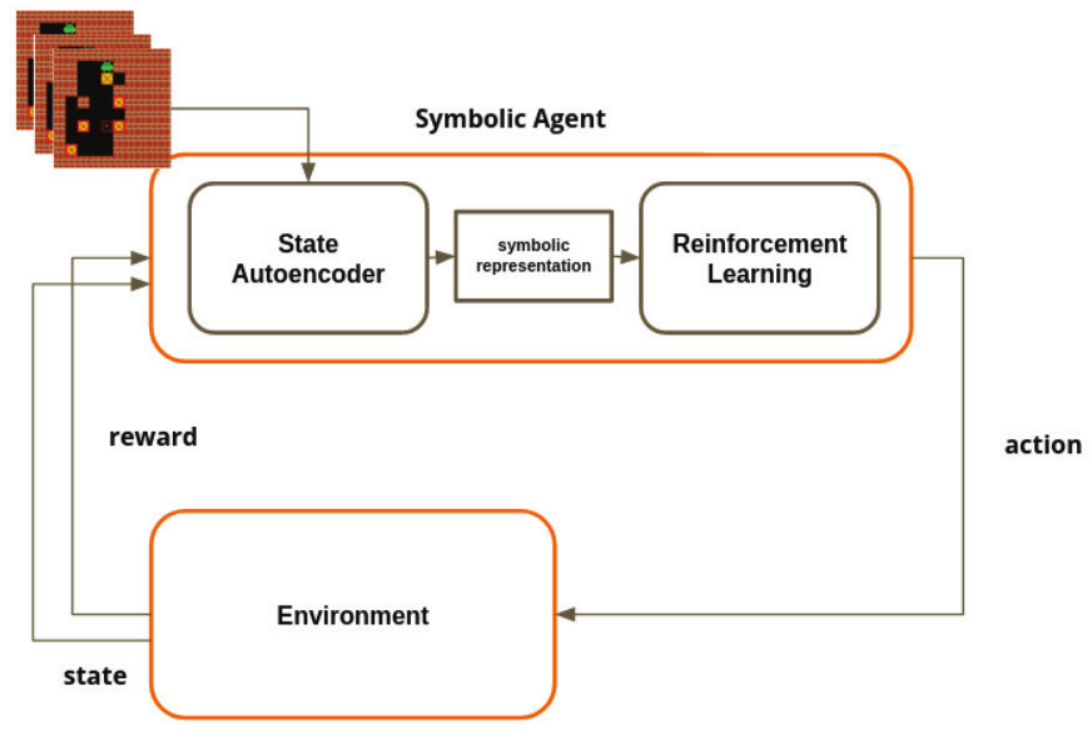} \vspace {-3mm}
		\caption{Grounding can be achieved by actively exploring the world and learning about the forms and meanings of entities the abstract digital symbols refer to. Reinforcement learning is an effective way to learn these symbols by interaction. Illustration courtesy \cite{symbolicRL2024}. 
		}\label{fig:11_Grounding_Interaction}
	\end{figure}
	
	\subsubsection{Grounding by active exploration and interaction with the environment}
	
	An important way to ground symbols is to actively explore the world to find the meanings of relevant entities \cite{macdorman1999grounding,jacobsson2007rule,kulick2013active}. Approaches for facilitating generalist capabilities of LLM agents by means of embodiment have been discussed in detail in Section 3. From the discussion, it is clear that the role of embodiment as a tool for knowledge acquisition is anchored in its ability to support deliberate actions or exploration and interactions with the world \cite{colle2024improving,roy2005semiotic,mecattaf2024little}. In addition to the cognitive skills that can be learned with the help of these interactions, the (embodiment) mechanism helps LLM agents to learn the meanings of abstract symbols through direct experiences with objects and phenomena in the world. Current approaches (e.g., \cite{hsu2024grounding,carta2023grounding,zadem2024llm,gehring2024rlef,tan2024true}) typically employ reinforcement learning techniques to directly connect language constructs, physical objects, abstract concepts and actions. To achieve this, the agent first learns to ground low-level symbols in tangible experiences. In turn, high-level concepts can be built on and grounded in these low-level symbols. Human-in-the-loop reinforcement learning approaches \cite{lilanguage,moroncelli2024integrating} have also been used to provide semantically richer grounding for high-level concepts. Since training agents with reinforcement learning requires a huge number of trials, virtual worlds are commonly used to simulate the behavior of the real world \cite{cao2024survey,hong20233d,guan2023leveraging,cheng2024llm+}. Figure \ref{fig:11_Grounding_Interaction} show a generalized architecture of this approach.

	\subsubsection{Leveraging external knowledge for LLM grounding}
	Besides the explicit symbolic grounding methods discussed, LLMs can also leverage external knowledge from diverse sources to provide “weak” grounding. So-called encyclopedic knowledge graphs \cite{pan2024unifying}, for example,  can represent a large volume of structured knowledge mined from diverse sources, including encyclopedia like Wikipedia \cite{vrandevcic2014wikidata} and relational databases \cite{mahria2023building,sun2020model}. Even though these approaches may not strictly involve abstract, primitive entities with connected by syntactic or logic rules, they still provide a bridge between the pure implicit knowledge in classical LLMs trained on large-scale generic datasets. Retrieval-Augmented Generation  (RAG)\cite{lewis2020retrieval,zhao2024retrieval,gao2023retrieval} is another common technique for grounding LLMs on external knowledge. The basic idea is to leverage additional information from external sources to augment available knowledge for the grounding process. RAG is particularly useful when there is the need to augment the generic knowledge stored with domain-specific knowledge within a very narrow context \cite{zhang2024raft,wang2024domainrag}. Another popular form of RAG, \textit{domain tool augmentation}, enables LLMs to access and use external tools and plugins through specially designed application programming interfaces (APIs) \cite{parisi2022talm,qin2023toolllm,panda2024revolutionizing}.

	\section{Causality}
	
	\subsection{Causality in artificial and human intelligence}
	Causality characterizes how various factors, phenomena or events influence other events, objects or processes in the real world \cite{scholkopf2022causality,pearl2009causality}. Causal learning in its simplest form is aimed at determining the dynamic relationships between two variables, where one variable, the cause, directly influences another variable, the effect. While grounding primarily deals with connecting primitive symbols such as words to their meaningful representations (i.e., variables, phenomena, concepts, etc.) in the physical world, causality is concerned with explaining the underlying mechanisms and reasons responsible for changes in these parameters and how they affect various outcomes in the world. In machine learning and AI, known casual relationships can be explicitly encoded by their human developers.
	
\begin{figure}[H]
		\vspace {-3mm}
		\centering
		\includegraphics[width=0.90 \linewidth]{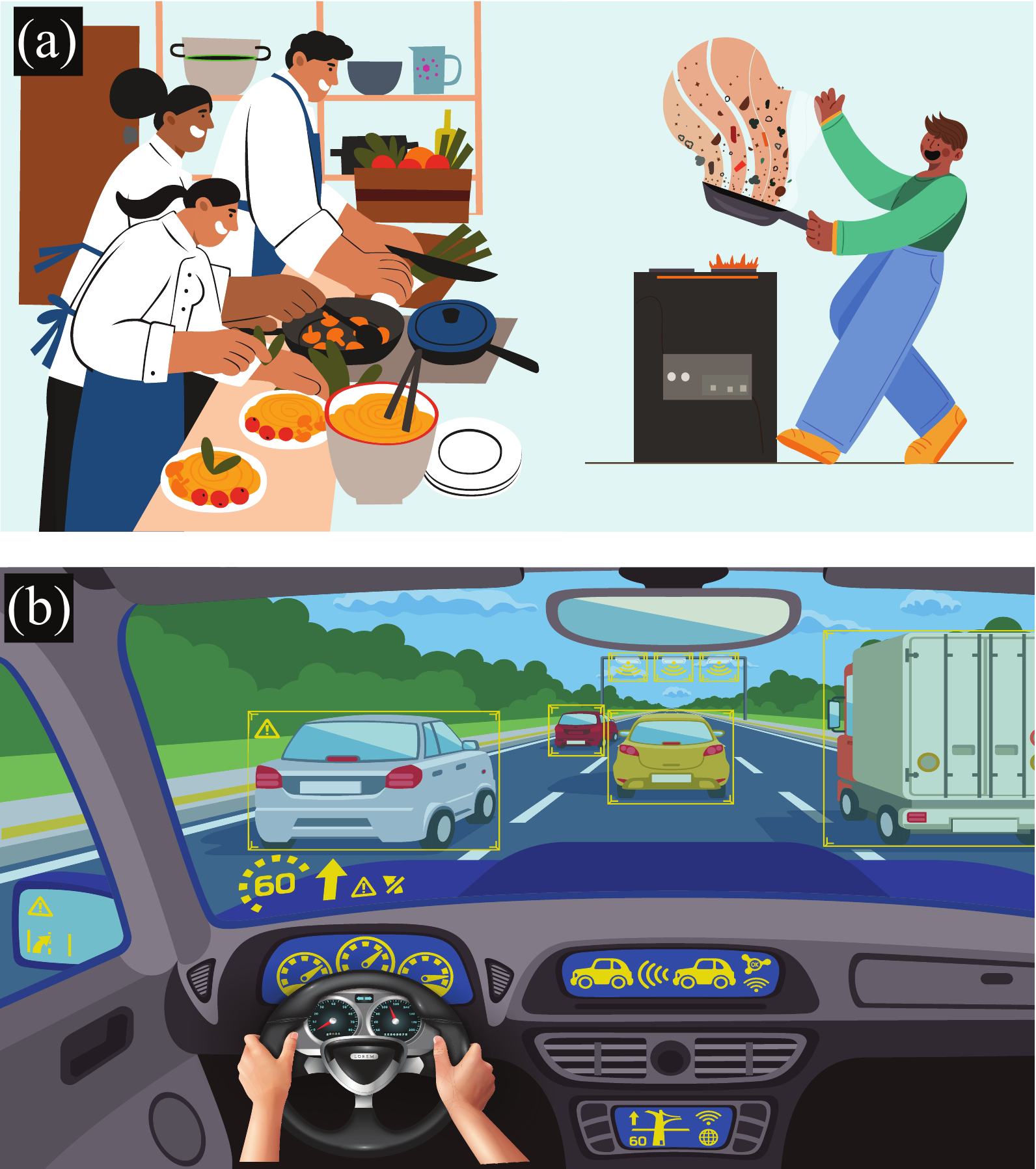} \vspace {-3mm}
		\caption{Interpreting events and observations or performing everyday activities such as cooking and driving requires an understanding of causal relationships. In cooking, for example (a), there is the need to understand concepts like volume, weight, boiling, as well as the behavior of entities such as fire, etc. Similarly, driving (b) requires an understanding of concepts such as speed, momentum, inertia, collision, and soo forth.
		}\label{fig:13_Activities_Causality}
	\end{figure}

	Causal reasoning – the process of leveraging the understanding of cause-and-effect relationships to explain events–enables AI systems to reason about (make accurate predictions about) complex real-world phenomena such as the resistance of structures to adverse weather elements, climate change, spread of diseases, accidents, population growth, economic performance, etc. This understanding is vital for everyday activities such as cooking, washing and driving (see Figure \ref{fig:13_Activities_Causality}).  Another vital role of causal reasoning is to improve robustness to interference and maintain correct inference when the underlying conditions and internal mechanisms or the environment changes. Thus, causal modeling allows AI systems to better generalize and transfer learned knowledge to new settings. Additionally, causality-aware models can account for inherent limitations and deficiencies of observations or data. For example, they can eliminate or mitigate the effects of adversarial examples \cite{zhang2021causaladv,ren2022towards,debbi2024causadv}. and biases \cite{vlontzos2022review}.

	\subsubsection{Basic principles of causality}
\begin{figure}[H]
		\vspace {-1mm}
		\centering
		\includegraphics[width=1.0 \linewidth]{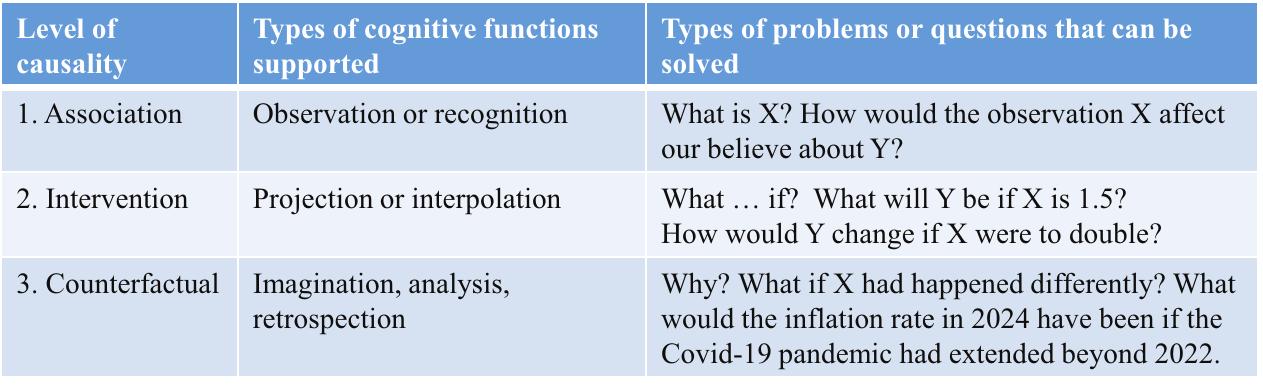} \vspace {-3mm}
		\caption{Levels of causality per Pearl \cite{pearl2019seven} and the types of problems each can handle.   
		}\label{fig:12_Levels_of_Causality}
	\end{figure}

	Causal understanding can be categorized into different degrees based on the level of causal reasoning power they permit. These degrees range from basic associations to reasoning about hypothetical scenarios. One of the most popular classification frameworks was formalized by Pearl in \cite{pearl2019seven}. It describes a three-level hierarchical scheme for classifying causal relations derived from observations, or more specifically, from data.  The levels are designated as Association (Level 1), Intervention (Level 2), and Counterfactual (Level 3) —see Figure \ref{fig:12_Levels_of_Causality} for a summary of the typical questions addressed by each of these levels of causality. According to this scheme, solving reasoning problems at any level is possible only if information from that level or higher is available.  The lowest level of causal reasoning, Association, relates to situations where answers to questions are obtained directly from observations in the form of statistical relations in the observed data. Intervention, the second level of causal reasoning, involves estimating the extent to which changing one variable (for example, treatment option) affects a target variable (i.e., a particular outcome, in this case recovery). Causal information at this level allows the effect of specific actions to be correctly predicted. For example, a force of 10 Newtons impacting a heavy truck would not cause any measurable motion. The third and highest level, Counterfactual, allows answering hypothetical questions or making inference about unobserved outcomes. This involves answering "what if this happened" type of questions- that is, what could have happened if certain events had not occurred or had happened differently. Counterfactual reasoning allows us to determine which variable to manipulate, and to what extent, in order to change a target variable to some desired state (obtain a desired treatment outcome). Solving intelligence problems at this level involves using both associative and interventive information. 
	

	\subsection{Approaches for modeling causality in LLMs}
	Learning causality (or causal modeling) generally aims to solve two problems:  (1) causal discovery – identifying the underlying mechanisms, their associated physical parameters and the interrelationships that govern the operation of the system; and (2) causal inference – the task of estimating the effects of causal variables on one another based on a pre-defined hypothesis about their causal relationships \cite{lobentanzer2024molecular}.  Causality modeling can be in the form of implicit learning of causal relationships \cite{willig2023probing} or explicit representation using prior knowledge about some domain-specific causal mechanisms and relationships \cite{hasan2022kcrl,constantinou2023impact}. Implicit causal learning methods rely on end-to-end deep learning methods to identify cause-and-effect relationships directly from data or apply this knowledge for inference. We discuss the important methods for modeling causality in the next subsections. A comparison of these methods is summarized in Table \ref{tab:approaches_causality}. 
	
	\subsubsection{Conventional deep learning methods}
	Multimodal LLMs trained on large-scale generic data have shown a great capacity to model causal relationships \cite{willig2023probing,kiciman2023causal}. This is mainly achieved by learning hidden patterns from the vast amount of training data. For instance, an LMM may be able to infer important variables that impact economic growth or inflation, as well as identify specific causal links between these variables via purely learned patterns, even when this information is not explicitly specified in the training data.  Knowledge acquired in this manner is constrained by the fact that not all observed connections have cause-and-effect relationships. Indeed, a large number of real-world phenomena exhibit correlative relationships \cite{marwala2015causality,stapp1976correlation} –a relationship in which changes in the target variables, by coincidence or by unrelated influences, follow each other but are actually not related by any causal link. In reasoning tasks such false correlations can lead to inaccurate or wrong conclusions. 
	
	In addition, owing to their ability to learn cause-effect relationships by discovering hidden patterns in training data, LLM’s are also trained on extensive text that describe cause-effect relationships, including mathematical relations, scientific principles and laws, etc. In the training process, for example, models can acquire knowledge about causal relations that allow them to handle high-level causal reasoning tasks– including interventions \cite{jin2023cladder,zhang2024if,zhang2022rock} and counterfactual reasoning \cite{feder2024causal,liu2023magic,zhang2024if,chen2022disco,zhang2023towards,li2023prompting} – from explicit statements such as “lack of physical exercise leads to obesity”.  Despite this seemingly powerful ability to model causality, researchers have shown that state-of-the-art LLMs are not able to acquire real causal reasoning capabilities \cite{zevcevic2023causal,koch2023babbling,wu2024causality}, even with additional training aimed at enabling causality, including in-context learning and finetuning \cite{jin2023can}. Specifically, LLMs trained purely on data, without internal causal modeling mechanisms, are, not inherently aware of physical laws or underlying mechanisms and principles that govern behavior in the real world, and their predictions are often merely based on learned correlations. This can lead to serious and dangerous mistakes. To address this limitation, researchers often have to rely on finetuning the models on specially-curated causal datasets (e.g., \cite{du2022care,dalal2023calm}) to discover causal relationships. However, this approach is a laborious and difficult task that is often impractical to scale up in complex real-world settings. Moreover, the method often requires many simplifying assumptions that can sometimes lead to incorrect relationships (see \cite{rawal2024causality,glymour2019review}).

	\subsubsection{Neuro-symbolic methods}
	In contrast to the DL approach which builds predictive model that mainly learn statistical dependencies, neuro-symbolic methods explicitly incorporate prior knowledge about causal mechanisms into the LLM model. One way to achieve this is by leveraging knowledge graphs \cite{ji2021survey,pan2024unifying} and other structured knowledge-based causal representation and inference techniques \cite{jonassen2013structural,pinto2019structured,shen2020exploiting}. As these models naturally encode relationships between concepts, many works (e.g., \cite{yang2023chatgpt,yu2024fusing,wang2024llmrg,susanti2024knowledge,ravi2024exploration,wang2024llmrg,abu2024knowledge}) utilize them to provide structured knowledge as causal mechanisms which the LMM then incorporates into its information generation process. 
	One of the most effective neuro-symbolic approaches to extending the causal reasoning ability of LLMs is to integrate causal graphical models \cite{elwert2013graphical,cheng2024data}, a special class of structured knowledge techniques, that, by their nature, are inherently causal. The basic approach is to represent causal assumptions formally using special diagrams or graphs. In the representation, nodes of a graph represent causal variables while edges indicate the existing causal relationships between the variables. By evaluating the effect of several variables, it becomes possible to determine whether or not the prior assumptions about causality are valid. And in the case where these assumptions hold, to derive mathematical expressions that describe the relationships. Conversely, the methods allow researchers to falsify causal assumptions. The idea is simple: to establish causation, isolate and induce a change of one of the possible factors. Where there is causation, it will manifest as a corresponding change in the target variable. The learned latent structural relationships are then integrated into the LLM’s neural network's learning process. A large number of works (e.g., \cite{li2021causalbert,ke2019learning,khetan2022causal}) that employ causal graphical models \cite{elwert2013graphical,cheng2024data} have demonstrated the potential of this . For instance, Wang et al. \cite{wang2024credes} propose a so-called Causal Relationship Enhancement (CRE) submodule that utilize Structural Causal Model (SCM) to model casual mechanisms for subsequent integration into an LMM framework. On the other hand, Samarajeewa et al. \cite{samarajeewa2024causal} employ external causal knowledge to augment LLM to improve causal reasoning. The authors argue that LLMs, though have shown strong reasoning capacity, still need additional causal knowledge from structured sources to adequately infer causal relations. To this end, they employ RAG technique to recover Causal Graphs as external knowledge sources to extend LLM’s causal reasoning capabilities. 
	
	Because of the tedious and time-consuming nature of the task of modeling causal mechanisms with the aforementioned graphical methods, some new approaches have been devised to leverage LLMs themselves to construct causal graphs models which, in turn, could be utilize to augment the LLMs. Since large language models themselves already possess extensive knowledge of real-world contexts, and patterns of behavior, including causal relationships between different variables \cite{li2024realtcd}, a large number of recent works (e.g., \cite{kiciman2023causal,ban2023query,liu2024discovery}) have proposed to leverage this knowledge to build causal graphs. In this line of work, the LLMs commonly serve as a source of prior knowledge about causality – i.e., to establish initial variables and dependencies \cite{darvariu2024large}– or as a means to augment already known causal relationships by suggesting additional causal variables \cite{sheth2024hypothesizing}. Typically, the LLM helps by describing the general structure of the graph in the form of variables (i.e., nodes) and their causal relationships (i.e., edges). With this method, there is also the possibility to interact with the LLM and exploit its reasoning abilities to refine the skeletal graph through prompting \cite{jin2023cladder,willig2023probing}.

	\begin{figure}[H]
		\vspace {-3mm}
		\centering
		\includegraphics[width=0.90 \linewidth]{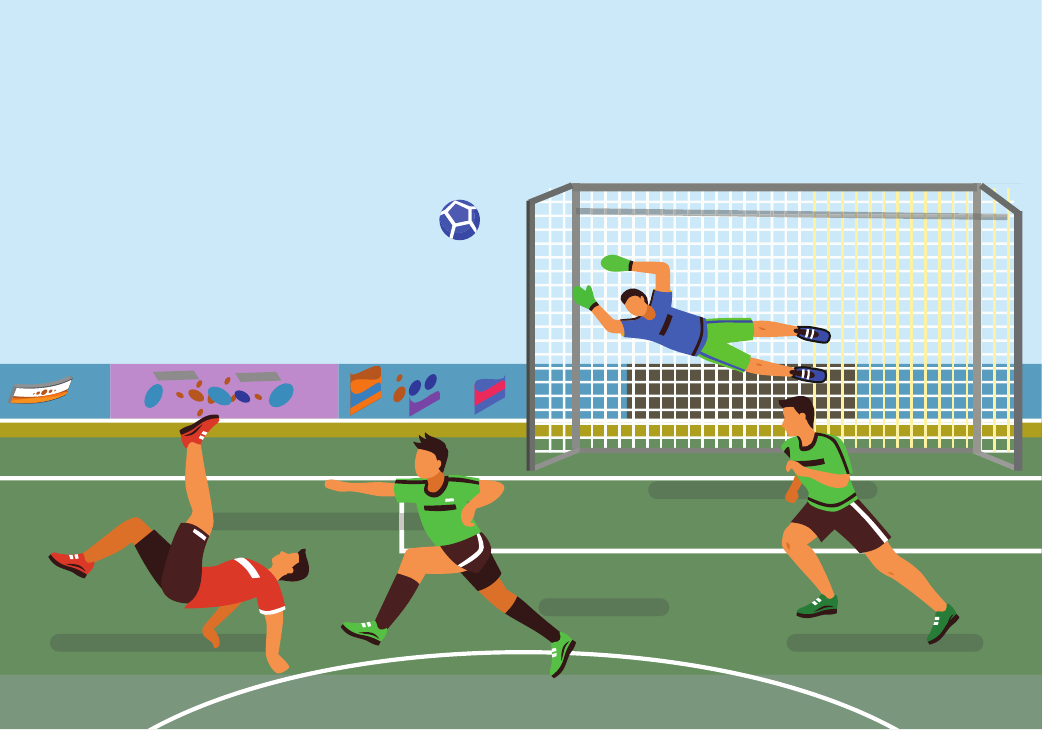} \vspace {-3mm}
		\caption{Unlike AI systems, humans naturally have an intuitive understanding of cause-and-effect relationships, including a rough knowledge about how physical properties of materials and systems affect their behavior. The player can roughly estimate the amount and direction of force needed to get the ball to the right location. Similarly, the goalkeeper has a rough idea about the ball's direction and speed based on the striker's movement and pose before the kick.
		}\label{fig:14_Human_CausalR}
	\end{figure}
	
	\subsubsection{Physics-informed world models}
	It has been hypothesized that humans’ ability to infer and reason about causal events relies on their world model (see, for example \cite{goldvarg2001naive,hafri2021perception,baradel2019cophy}).  This world model, or metal model \cite{jones2011mental}, encode causal abstractions of concepts, phenomena and objects in the world in a way that maintains a definite, albeit fuzzy, structure and rules governing behavior.. Based on these abstractions, humans have rough, implicit knowledge about the world in the form of intuitive physics \cite{kubricht2017intuitive}– i.e., basic properties of various entities and how these properties influence behavior. This allows humans to make unconscious but quick judgments about the physical interactions in the environment, for example, they can make fairly precise judgements about how objects move, fall, or collide (Figure \ref{fig:14_Human_CausalR}).

	\begin{table*}[h!]
	
	\caption{A comparison of the strength of causality provided by the different classes of approaches. Note: In this table, w.r.t. means with respect to; Assoc. refers to the association level of causality; Interv. is the intervention level; and Count. is the counterfactual level.}
	\label{tab:approaches_causality}
	
	\scalebox{0.85}{

			\begin{tabular}{@{}llllll@{}}
				\toprule
				& \multicolumn{3}{l}{\textbf{Capability w.r.t. levels of causality}} & \multicolumn{2}{l}{\textbf{General comparison}}                                                                                                                                                                                                                                                                                                               \\ \midrule
				\textbf{Class of approach}                                                                                & \textbf{Assoc.}      & \textbf{Interv.}      & \textbf{Coun.}      & \textbf{Strengths}                                                                                                                                                           & \textbf{Weaknesses}                                                                                                                                                            \\
				\begin{tabular}[c]{@{}l@{}}Deep learning  \\ (e.g., \cite{du2022care,dalal2023calm})\end{tabular}          & Yes                  & No                    & No                  & \begin{tabular}[c]{@{}l@{}}Does not require prior domain \\ knowledge; easy to scale; simple\end{tabular}                                                                    & \begin{tabular}[c]{@{}l@{}}Can only establish  correlations, \\ requires specific datasets for causality.\\ Prone to serious errors and catastrophic \\ failures.\end{tabular} \\   \\
				\begin{tabular}[c]{@{}l@{}}Knowledge graphs \\ (e.g., \cite{ji2021survey,pan2024unifying})\end{tabular}   & Yes                  & Partial                & No                  & \begin{tabular}[c]{@{}l@{}}Can model complex relationships,\\ can readily  be integrated into deep \\ learning frameworks\end{tabular}                                       & \begin{tabular}[c]{@{}l@{}}Not all data types are supported, modeling \\ casual relationships  can be laborious and \\ time-consuming; difficult to scale.\end{tabular}        \\    \\
				\begin{tabular}[c]{@{}l@{}}SCM  \\ (e.g., \cite{elwert2013graphical,cheng2024data})\end{tabular}          & Yes                  & Yes                   & Yes                 & \begin{tabular}[c]{@{}l@{}}Can model causal  dependencies in \\ a comprehensive way; can be used \\ for interventions and counterfactual \\ reasoning\end{tabular}           & \begin{tabular}[c]{@{}l@{}}Difficult to implement,   requires complete \\ knowledge about all factors, extremely\\ difficult  to scale.\end{tabular}                           \\  \\
				\begin{tabular}[c]{@{}l@{}}Physics models  \\ (e.g.,, \cite{cherian2024llmphy,zhaobridging})\end{tabular} & Yes                  & Yes                   & Yes                 & \begin{tabular}[c]{@{}l@{}}Extremely scalable,   models can\\ leverage existing world models, \\ can be used for interventions and  \\ counterfactual reasoning\end{tabular} & \begin{tabular}[c]{@{}l@{}}Require enormous  computational resources, \\ requires knowledge of factors and their\\ relationships.\end{tabular}                                 \\ \bottomrule
			\end{tabular}
				
} 
\end{table*}
	
	In line with this idea, many recent works \cite{duan2022survey,ahmed2020causalworld,baradel2019cophy} leverage virtual worlds based on intuitive physics engines \cite{cherian2024llmphy,xue2024phyt2v,zhaobridging} to ground LLMs’ knowledge in causal properties and behavior of the real world. These models employ formal, mathematical models designed on the basis of prior knowledge to represent physical laws about the world. As a result, they simulate causal relationships and effects of real-world phenomena such as aerodynamics, gravity, force, and lighting, and heating. LLM-based AI agents that interact in such worlds in the process of training learn generalizable causal laws and behavior (e.g., flying, falling, burning, deformation, floating on water and breaking into pieces, etc.). Using knowledge from areas such as psychology and anthropology, it is also possible to model human behavior as well as social interactions \cite{gui2023challenge,gurcan2024llm}. This allows “common sense” reasoning about observations and interactions. The inherent relations allow agents to easily handle questions about counterfactual. Agents themselves can be intrinsically designed as embodied virtual models capable of seamlessly interacting with the simulated causal world. Such agent models typically incorporate external frameworks or submodules that utilize accurate mathematical relations to model physical laws that describe the agents’ own properties and behavior. This way, embodied agents such as robots can predict the impact of their own actions as well as various physical influences on themselves. 
	Causal modeling approaches that utilize virtual models have a number of strengths. They allow to simplify the complex process of obtaining large corpora of real data for training. Importantly, the underlying mathematical relations about physical interactions are invariably grounded in rigorous Newtonian physics, thermodynamics, or – depending on the level of realism required –particle physics and quantum mechanics. Therefore, the causal knowledge encoded in this class of models is precise and is often without any ambiguity, allowing exact results of interactions to be defined. Because of this representational power and precision, causal models based on physics engines can handle complex phenomena with consistency and accuracy far beyond human intuitive understanding. Despite these advantages, virtual models impose a number of limitations on the range and complexity of skills that can be learned.
	
	One of the main difficulties with this line of work is that highly detailed, large-scale simulations often require significant processing power. It is also enormously challenging to model such detailed, highly precise physical relations. However, the performance of the resulting AI systems is limited by the quality and completeness of the models used. Simulations may fail to accurately account for fuzzy concepts or factors unknown to the human developers. For example, simulations about human interactions may not accurately account for cultural, social and emotional factors. To overcome some of these challenges, approaches to incorporate deeply learned knowledge have been proposed. These include learning intuitive physics from data using special deep learning techniques \cite{piloto2022intuitive,hao2022physics} or learning interaction policies using reinforcement learning with human feedback \cite{wang2024srlm,chaudhari2024rlhf,laleh2024survey}.  Another common limitation is that the interactions modeled by these techniques are rigid and operate strictly according to the physical parameters and rules encoded. To mitigate this drawback, some works \cite{chan2024balancing,tremblay2018training} have proposed to leverage data-driven optimization techniques to induce a degree of variability and randomness in physics-based models.

	\section{Memory }
	
	\subsection{Basic concept of memory in biological and AI systems }
	Memory mechanism differs fundamentally from the other cognitive processes covered in Sections 2 through 4– embodiment, grounding and causality –in the sense that it primarily serves as a means to preserve, consolidate, and subsequently make available the important knowledge acquired through these other processes. The memory mechanism does not lead to the generation of fundamentally new knowledge about the world, but mostly restructures the knowledge that has already been obtained. Thus, the main role of the memory mechanism in AGI is to reconstruct and organize the already acquired knowledge for high-level cognitive tasks and for storage for future reuse. When new knowledge emerges in the process, it is often as a result of this reorganization process \cite{versace2014act}. Memory facilitates continual \cite{lavin2021simulation,guo2020improved} or lifelong learning \cite{bang2021rainbow}, an important feature of biological intelligence \cite{kudithipudi2022biological}. Memory can also serve as a means to incorporate prior knowledge into the AI system \cite{zhang2024memsim,anokhin2024arigraph}. 
	Living organisms of the same species often jointly occupy a given ecological niche and constantly interact with one another, mostly in a cooperative way. For this reason, an important aspect of their intelligence depends on the ability to learn and maintain knowledge about shared behavioral characteristics that govern their interactions. Humans, in particular, commonly rely on shared structured knowledge in the form of norms, rules, belief systems and customs that allow them to interact seamlessly in social settings \cite{adams2012beyond}. In addition to memorized information, humans and other higher living organisms have in-built prior knowledge or innate knowledge (see 
	\cite{taatgen2017cognitive,griffiths2009distinction}) that is encoded in genes and transmitted from parents to offsprings. To realize a similar function, AI approaches typically incorporate structured knowledge related to specific tasks in the form of knowledge graphs (e.g., in \cite{feng2023knowledge,li2024kgmistral}) or ontologies (e.g., \cite{bendiken2024know,allemang2024increasing}) to augment learned knowledge stored in memory. They comprise not only facts and properties of specific concepts or objects about the world, but also relationships and general rules about the world. Together with the learned knowledge, allow effective and meaningful inferences about newly encountered situations. 
	
	Besides storing and retrieving information, the memory mechanism serves as a means to bypass costly computations by reusing already computed cognitive variables and solutions \cite{dasgupta2021memory,bordalo2023memory}. For instance, when humans first learn a new task such as driving, it takes constant attention and conscious effort to perform it. However, after ingraining the requisite skill in memory through constant practice, the learned tasks can be performed effortlessly without much attention.  The phenomenon is well-grounded and supported by evidence from psychology. This saves scarce cognitive resources for new skills and conserves energy. Skills for mentally-engaging cognitive tasks in domains such as mathematics and complex games that require analytical reasoning particularly benefit from this phenomenon (for details about this see \cite{dasgupta2021memory,bordalo2023memory}). Memory has also been shown to play a key role in metacognitive tasks, where existing knowledge about a domain facilitates learning new skills \cite{ganapini2022thinking,spivack2024cognition,langdon2022meta}.  In addition, phenomena like imagination and mental imagery also illustrate the reuse of previous computations to facilitate efficiency \cite{dasgupta2021memory}.

	\subsection{General approaches to realizing memory in LLMs }
	The main techniques for implementing memory in LLMs are:

	\begin{itemize}
		
		\item Parameters in deep neural networks
		
		\item Attention mechanism
		
		\item Explicit memory
		
		\item Adequate diversity and variability
		
		\item External memory (e.g., through RAG)
	\end{itemize}
	
	\subsubsection{Memory as model parameters}
	Classical deep learning approaches store task-relevant knowledge as model parameters. Techniques such as finetuning and in-context learning seek to incorporate new knowledge by modifying these learned parameters instead of requiring information to be stored in independent, explicit memory. This often alters the model parameters and invariably leads to the loss of important information, a phenomenon commonly known as catastrophic forgetting \cite{goodfellow2013empirical}. A common workaround is to freeze some model parameters during the finetuning process to ensure that only knowledge that needs to be modified is affected \cite{ziegler2019fine,aghajanyan2020better}. More recent techniques include elastic weight consolidation (EWC) \cite{kirkpatrick2017overcoming}, unsupervised replay \cite{tadros2022sleep} and adversarial neural pruning \cite{figueroa2024evaluating}. Using knowledge editing techniques \cite{wang2024deepedit,zhong2023mquake,zhang2024comprehensive}, it is also possible to directly modified the learned knowledge instead retraining the model by the finetuning approach. 
	
	\subsubsection{Attention mechanism}
	Another way to obtain memory in LLMs, or in neural networks in general, implementations is to utilize attention mechanism to temporarily hold and process information from past input sequence.  While most modern LLMs are based on transformers, earlier language models employed various variants of recurrent architectures, including recurrent neural networks (RNNs) \cite{schmidt2019recurrent,mnih2014recurrent}, long short-term memory (LSTM) \cite{graves2012long,van2020review},  gated recurrent units (GRUs) \cite{dey2017gate} that explicitly capture and retained fragments of previous inputs through hidden states using attention mechanism. In essence, the attention mechanism in this case provides short-term memory that allows the model to “remember” recent sequences or, more technically, to maintain context within the given sequence. However, the memory capacity in this case is very limited. 
	Some works (e.g., \cite{hatalis2023memory}) employ the context window of the LLMs as memory, where information contained in prompts leveraged as state, task or goal description. This information is treated as working memory. The information in the context window can also be in the form of high-level concepts in natural language. These may be object or environment properties, task goals, desirable skills or attributes of the agent itself. Owing to the restricted memory capacity of the LLM context window, the volume of information that can be handled by this type of memory mechanism is very small. Moreover, recent work (e.g., \cite{peysakhovich2023attention,hsieh2024found,liu2024lost})  has shown that models often exhibit bias for information at the beginning and end of the context window, prioritizing these portions while ignoring the middle. Consequently, very long contexts may result in a situation where a large portion of information (outside the two extremities are not memorized, culminating in the so-called Lost in the middle problem \cite{liu2024lost}.  Because of these limitations, explicit memory has been proposed as a viable workaround that allows practically unlimited memory capacity for large-scale knowledge storage for generalist agents.  
	
	\subsubsection{Explicit memory}	 
	To address the shortcomings of aforementioned memory techniques, approaches have been devised to allow selective storage of persistent, task-relevant information in computer memory for reuse \cite{hatalis2023memory,maharana2024evaluating,hou2024my,wang2024karma}.  In particular, domain-specific knowledge can be stored explicitly in memory as prior knowledge to augment the extensive generic knowledge learned by the LLM. Learned knowledge in the form of agent experiences (i.e., past decisions, actions or attempted actions and feedback from the environment) may also be stored explicitly in memory\cite{yin2024explicit,wu2024longmemeval,zhang2024survey,packer2023memgpt}.  The essence of this memory system is to sample and accumulate useful experiences over time in the process of interaction with the environment.
	Relational databases \cite{hu2023chatdb,liu2023think,hatalis2023memory,qin2024relational} are one of the most popular type of storage for traditional information-intensive tasks. The stored information can then readily be retrieved using sequential query language (SQL) queries. This approach also allows information to be readily saved on external database servers and then retrieved as needed. Since the data format of traditional relational databases is often not designed to be readily usable by LLMs, a common workaround is to utilize structured databases \cite{jiang2023structgpt,patel2024lotus} for knowledge storage. Vector databases \cite{wen2023dilu,zhang2023long,jing2024large,han2023comprehensive}, in particular, are very useful for this purpose. This type of memory system can store specific facts, concept definitions and entity relationships in the form of knowledge graphs, which the LLM can query to aid inference. Memory mechanism based on vector databases allows not only for quick retrieval, but also permits sophisticated and fine-grained manipulations to be carried out at the feature level (i.e., feature vector space). Moreover, this representation method makes it easy to use learned operations acquired in the training phase instead of relying on predefined analytical routines to manipulate the stored information. 
	
	\subsubsection{External memory through RAG}
	With regard to generalist capabilities, one of the key advantages humans have over other animals is the ability to use external knowledge resources and tools to augment or extend their competences (e.,g., by reading manuals, books, or by browsing the internet for the necessary information for a given task). This alleviates the need for storing all needed knowledge internally. Motivated by this prospect, recent works have sought to enable LLMs to access and utilize external resources, thus, extending the range of tasks they can perform. This also helps to overcome inherent limitations associated with insufficient memory and processing power. For example, using retrieval-augmented generation methods \cite{zhao2024retrieval,schmied2024retrieval,wu2024retrieval,li2022survey}, models can also query external knowledgebases to retrieve additional information when the they do not find the required knowledge locally. This information from external sources can be processed and utilized straightaway or can be stored in local memory for later use. Large volumes of knowledge can be stored for long time spans since the volume of information that can be stored does not depend on the agent’s memory capacity. There is also the ability to exploit rich and diverse knowledge resources already available (e.g., web portals, wikis, etc.). Besides shortening development time and simplifying the development process, the ability to exploit readily-available external knowledge also offers an economically cheaper way to realize advanced capabilities. The main drawback of this approach is that the external information may not be guaranteed to work correctly as a result of unknown errors, including the presence of unknown errors, inconsistencies or incomplete information. There can also be a complete loss of access to the external information for various reasons, including a change in access privileges, loss of storage resource or the information itself. The external information may also be more exposed to other users and malicious actors, thereby endangering security.
	
	\subsection{Memory types and their characteristics, role and implementation in LLMs}
\begin{figure*}[!htb]
		\vspace {-1mm}
		\centering
		\includegraphics[width=0.80 \linewidth]{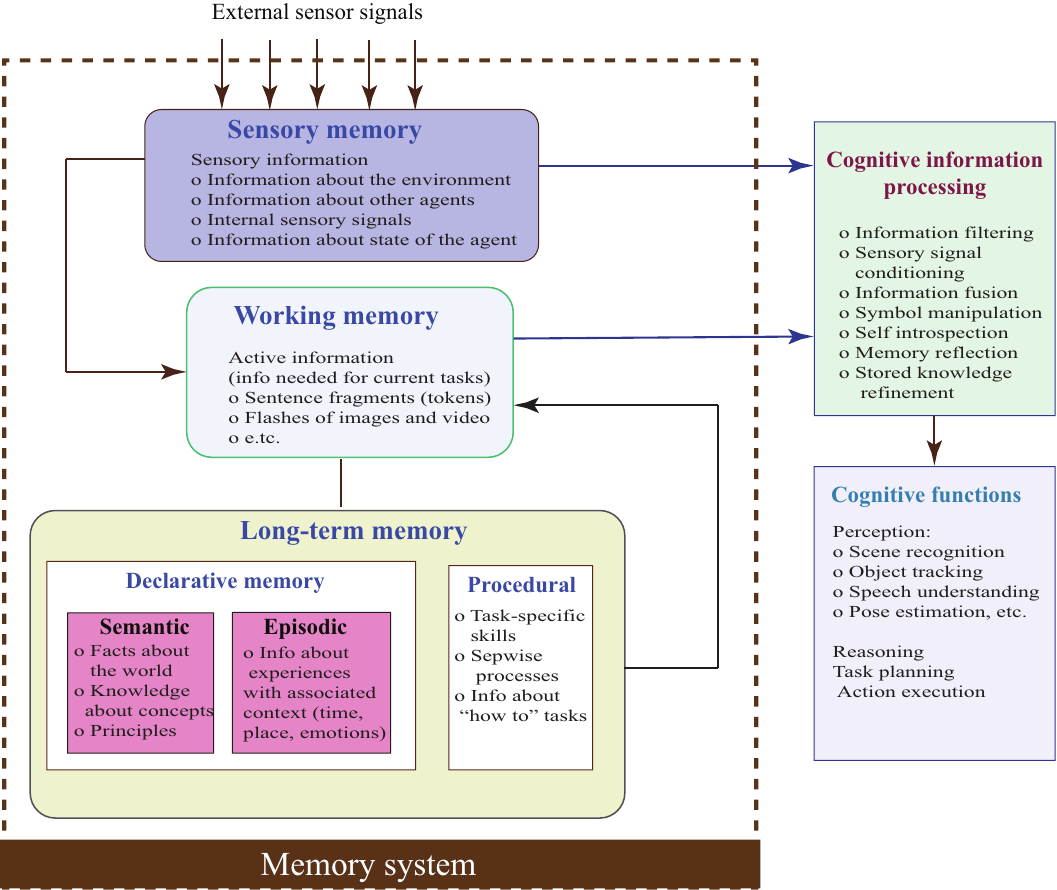} \vspace {-3mm}
		\caption{A simplified representation of the memory system showing information flow as well as the interaction various components and the cognitive system.  
		}\label{fig:15_Memory}
	\end{figure*}

	Three types of memory systems are commonly identified: sensory memory (SM); working memory (WM), also referred to as short-term memory (STM); and long-term memory (LTM). We discuss the general characteristics, main functions, role and methods of implementation of these various memory types in LLMs. A generalized memory structure of a typical cognitive system is shown in Figure \ref{fig:15_Memory}. The main functions and approaches for implemention of each of these forms of memory are summariszed in Table \ref{tab:T3_Summary_of_Memory_types}.
	
	\subsubsection{Sensory memory}
	Sensory memory mechanism is the initial stage in the information processing pipeline and it is primarily associated with registering sensations.  The role of sensory memory is to record perceptual inputs from various sensors and input systems (e.g., text, joystick and other control inputs) of the agent. Additionally, by focusing on more salient information while ignoring noisy signals, sensory memory serves as a filter for the vast amount of continuous sensory information from the environment. In terms of storage duration, this type of memory preserves information for the shortest amount of time. It retains information only briefly for sensory systems to access. That is, acting as a buffer, it makes sensations persistent enough to overcome the inherent inertia associated with sensory processing systems.  In biological cognitive systems, sensory memory is automatic and not under voluntary control. Similarly, sensory memory systems in artificial intelligence systems can be implemented as a form of latching systems that buffers input signals \cite{ali2024robots,minnella2023mix}. This function– i.e., holding sensory inputs for finite amount of time – allows the cognitive system to produce continuous perception of the environment. This “perceptual continuity” (creates a persistent experience of reality, which) is useful for understanding and interacting with the world in a coherent manner. 
	Memory buffer can be realized natively by leveraging latches and buffered data mechanisms typically employed in microprocessor systems dedicated to real-time signal processing \cite{minnella2023mix,meaney2024synchronous}.  Peripheral interfaces for such microprocessor systems may utilize a kind of read buffer mechanism in order to avoid read misses as a result of delayed access.  This buffered peripherals usually employs a dedicated register or set of registers that holds the last data received from the external peripheral \cite{grotschla2023design,leis2023virtual}. This data is typically erased only when new data is written to the buffer or when an explicit deletion is requested by software. Write-back buffer can also be used at the destination to allow time for the processing elements to access sensory data. Such a buffer mechanism helps to ensure continuous access and availability of the sensory information during processing.
	
	Bio-inspired techniques that imitate the sensory memory mechanism of biological neural systems have also been proposed (e.g., in \cite{wan2020artificial,ji2023brain}).  These approaches are commonly based on advanced material technologies—technologies that produce artificial systems that exhibit useful properties of biological sensory systems. 
	
	Even though most state-of-the-art large language models do not explicitly incorporate sensory memory, the working principle as described in the preceding paragraph is naturally implemented in cyber-physical systems that utilize sensors for perception. Thus, with this loose definition of sensory memory mechanism, one can assert that almost all embodied LLM systems that read sensory signals implement sensory  mechanism of some sort. 
	
	\subsubsection{Working memory}
	Working memory \cite{lu2023memochat,park2023generative,liu2023think,wang2023voyager}, also known as short-term memory, retains relatively small amounts of information in an active, readily accessible state for relatively short durations during/for processing for immediate cognitive tasks, including perception, decision-making, reasoning, instruction following and executive functions (i.e., sensorimotor control). The biological concept of short-term memory was first described by NC Waugh and DA Norman in \cite{waugh1965primary}. Baddeley \cite{baddeley1992working} later introduced the concept of working memory, a model that conceptualizes short-term memory as a block of memory in which information needed for the execution of current tasks circulates. Thus, the content of short-term memory is retained until the target task is completed, after which the information is either forgotten or is saved to long-term memory. 
	
	High-level contextual information about interactions is also maintained in working memory to facilitate more complex tasks such as abstract reasoning. For example, in reinforcement learning-based agents, real-time information relating to agent trials and feedback are commonly maintained in working memory to facilitate reasoning \cite{shinn2024reflexion}. By momentarily holding recent inputs for immediate processing, the context window in LLMs can be regarded as serving analogous function as working memory. The LLM typically holds the most recent tokens in the context window, remembering only what was presented a few sentences away from the present. The content, similar to working memory, easily decays or overwritten by new input. Thus, it functions as the “memory” from which the model can retrieve and process recently inputted tokens before they are discarded or replaced by new inputs. The size of the context window determines the capacity of this memory. Indeed, many works precisely utilize the LLM context window as short-term memory mechanism \cite{hatalis2023memory,packer2023memgpt,chen2023walking}. In this regard, the information in the context window is leveraged as intermediate task, state or goal description. LTM content can also be utilized/retrieved by prompts to extent/enrich the information represented in the context window. Conversely, information in the context window can also provide additional knowledge to enrich long-term memory \cite{li2024needlebench,xiao2024infllm}. 
	
	\subsubsection{Long-term memory}
	Long-term memory preserves information for long time spans. The content of long-term memory is information selected from working memory that is deemed useful for long-term  storage\cite{chang2011short,baddeley1988long,ericsson1995long}. This ensures that the most relevant and important knowledge is available for reuse. In biological cognitive systems, long-term memory can hold information for the entire lifetime of the organism \cite{bekinschtein2008bdnf,park1996mediators}, though over time the information may be subject to decay, distortion, or loss \cite{hardt2013decay,winocur2010memory}. The problems of information loss can be avoided in synthetic memory systems by incorporating explicit schemes that maintain information permanently \cite{modarressi2024memllm,yamada2023developing}. With effectively unlimited capacity, long-term memory is the ultimate reservoir of knowledge, accumulated experiences and skills that can be recalled and utilized when needed. There are two main types of long-term memory systems commonly implemented in LLMs: declarative and procedural memories \cite{ullman2001neurocognitive}.

	\begin{table*}[h!]
		\caption{A summary of the main memory types and the common approaches for their realization. The asterisk (*) symbol Indicates long-term memory.}
		\label{tab:T3_Summary_of_Memory_types}
		
			\scalebox{0.85}{
	
			\begin{tabular}{@{}llll@{}}
				\toprule
				\textbf{Type}                                                & \textbf{Main functions}                                                                                                                                                                                                                                                                                                                                                           & \textbf{Methods fo realization in   LLMs}                                                                                                                                                                                                                                                                                                       & \textbf{Rep. works}                                                  \\ \midrule
				\begin{tabular}[c]{@{}l@{}}Sensory\\ memory\end{tabular}     & \begin{tabular}[c]{@{}l@{}}Stores sensory information momentarily for the\\ cognitive system to access.\\ Servers as a buffer mechanism to allow cognitive\\ processing components to work at different speed\\ without information lost.\end{tabular}                                                                                                                            & \begin{tabular}[c]{@{}l@{}}Buffered I/O systems; latching mechanisms for\\ sensory inputs;  bio-inspired sensory   memory\\ techniques .\end{tabular}                                                                                                                                                                                           & \cite{ali2024robots,ji2023brain,wan2020artificial}                   \\  \\
				\begin{tabular}[c]{@{}l@{}}Working\\ memory\end{tabular}     & \begin{tabular}[c]{@{}l@{}}Filters information (e.g., through attention mech-\\ anism) for long-term storage.\\  Holds active information temporally for cognitive\\ processing\end{tabular}                                                                                                                                                                                      & \begin{tabular}[c]{@{}l@{}}Attention  layers  in  deep learning  networks; \\ LMM context window.\end{tabular}                                                                                                                                                                                                                                  & \cite{hatalis2023memory,packer2023memgpt,chen2023walking}            \\  \\
				\begin{tabular}[c]{@{}l@{}}Semantic\\ memory*\end{tabular}   & \begin{tabular}[c]{@{}l@{}}Maintains and   accumulates  information  about\\ general rules, facts, principles,  relationships  of \\ various entities and general knowledge about the \\ world. \\ Allows prior knowledge about the world to be \\ reused in tacking cognitive problems tasks.\end{tabular}                                                                       & \begin{tabular}[c]{@{}l@{}}Structured knowledge in graphs; embeddings in\\ deep layers of neural networks; vector databases;\\ external knowledge accessed using methods such\\ as retrieval-augmented generation; in-context   \\ learning of semantics.   \\    \\ .\end{tabular}                                                             & \cite{anokhin2024arigraph,gutierrez2024hipporag,geva2020transformer} \\ \\
				\begin{tabular}[c]{@{}l@{}}Episodic\\ memory*\end{tabular}   & \begin{tabular}[c]{@{}l@{}}Integrates multimodal information (spatial, audio,\\ auditory, visual, olfactory, etc.) to form a   unified, \\ contextually-grounded representation that captures\\ “personal” experiences.\\  Allows agents to introspect   and reflect on the past\\ in the context of current reality, and, thus, provides\\ a means to self-improve.\end{tabular} & \begin{tabular}[c]{@{}l@{}}Explicit storage of specific events and experiences\\ with the associated temporal, spatial, social, etc. \\ contexts; in-context learning of significant events\\ by active prompting;\end{tabular}                                                                                                                 & \cite{fountas2024human,barmann2024episodic}                          \\  \\ \\
				\begin{tabular}[c]{@{}l@{}}Procedural\\ memory*\end{tabular} & \begin{tabular}[c]{@{}l@{}}Stores information about the detailed steps involved\\ in performing various specific (often, sensorimotor)\\ activities. \\ Allows agents to perform complex activities without\\ cognitive computations. This automates tasks, boots\\ speed and saves cognitive resources for other funct-\\ ions.\end{tabular}                                     & \begin{tabular}[c]{@{}l@{}}Explicit, long-term storage of real-time information\\ obtained from interactions with the  environment;\\ explicit encoding of task plans using neurosymbolic\\ methods; external knowledge  about task- specific\\ action sequences (e.g., using knowledge graphs);\\ retrieval-augmented generation.\end{tabular} & \cite{aktas2024vq,silver2022pddl}.                                   \\
				&                                                                                                                                                                                                                                                                                                                                                                                   &                                                                                                                                                                                                                                                                                                                                                 &                                                                      \\ \bottomrule
			\end{tabular}
	
	} 
	\end{table*}
	
	\textbf{(a) Declarative memory}
	Declarative memory is also known as explicit memory in biological cognition as the content can be consciously interrogated and recalled \cite{eichenbaum1997declarative,eichenbaum1999hippocampus}. In the context of artificial intelligence, declarative memory involves knowledge about specific facts that can be explicitly represented and retrieved \cite{wang2024memory}. Declarative memory is further subdivided into semantic and episodic memories \cite{tulving1998episodic}.  
	
	\textit{Semantic memory}
	Semantic memory maintains general knowledge that does not depend on specific contexts or agent’s “personal” or unique experiences.  The knowledge encoded in semantic memory consists of facts, formulas, general rules and laws, definitions as well as words and symbols and their meanings. The semantic memory focuses on high-level, conceptual knowledge about the world and how these are expressed in terms of symbols (e.g., words), graphics and speech (audio). In addition to facts about the world, semantic memory allows general rules and abstract principles to be retained for later use.  These rules are leveraged to manipulate new information in cognitive information processing. In this regard, they serve as a reasoning framework for interpreting the world as well as for acquiring and evaluating new knowledge.
	
	In large language models, semantic knowledge is naturally captured during training. In the course of training, multimodal language models learn to associate words, phrases, images, symbols and concepts based on the statistical patterns in the training data. This process allows the LLM to build a rich internal representation of general knowledge, which is maintained in long-term memory and recalled in the future to support cognitive tasks. By virtue of this knowledge, generic LLMs are impressive in reasoning tasks \cite{kojima2022large,xiao2024logicvista,li2024llms} and answering questions that require factual information \cite{shao2023prompting,zheng2024systematic}. For instance, questions like "What is the largest city in California?" or "How many feet make one kilometer?" are easily handled by these models. They are also adept at mining general rules from data and applying them in new tasks (see \cite{saeed2021rulebert,zhou2024wall,sun2024beyond}). This property of LLMs is the foundation of their impressive commonsense and analytical reasoning capabilities \cite{hu2024can,imani2023mathprompter}.
	Semantic memory can also be implemented in the form of prior knowledge encoded in structured forms such as knowledge graphs and causal graphical models can serve as long-term memory in LLMs. These structured knowledge submodules within LLMs can store facts, rules, concepts and relationships in a persistent manner, thus, allowing LLMs to retrieve useful information as needed. Works such as AriGraph \cite{anokhin2024arigraph}, HippoRAG \cite{gutierrez2024hipporag} and KG-Agen \cite{jiang2024kg} specifically employ structured knowledge forms as long-term memory. These specialized representation frameworks can particularly simulate the complex structure and interrelationships of various entities that is intended to be capture by information in semantic memory \cite{anokhin2024arigraph}. In this representation, where high-level concepts, their properties and relationships are explicitly connected.


	\textit{Episodic memory}
	Episodic memory maintains information about important events, experiences and associated contextual information \cite{tulving1972episodic} – that is, information about the times, locations, surrounding background or situational context, and nature of the events (e.g., the visual imagery, specific characteristics, including taste, touch, sound and other sensory signals that accompanied the event). This information is represented as time-ordered sequence of experiences. The information is not additive, that is, different events recorded are separate and the experiences are not generalized  nor accumulated. 
	Episodic memory mechanism in large language models can be realized in several ways, including by utilizing the attention mechanism of the underlying transformer architecture to capture episodic knowledge; transferring relevant episodic information from the LLM’s context window to long-term storage; through fine-tuning generic, pre-trained LLM frameworks on specific datasets that explicitly contain episodic knowledge which can be saved to long-term memory and recalled in the future.

	\textbf{(b) Procedural memory}
	Procedural memory deals with the acquisition, storage, and recall of knowledge about the logical steps needed to carry out complex activities. This commonly involves motor skills such as autonomous driving, cooking and robot manipulation. In the realm of large language models, procedural memory facilitates cognitive functions such as activity planning, instruction following, reasoning and execution of physical actions. This set of high-level cognitive capabilities are a core part of LLMs’ emergent abilities.

	In biological cognitive systems, a major aspect of procedural memory involves subconscious processes \cite{schacter1987implicit} – processes occurring outside the agent’s awareness, e.g., priming \cite{schacter1992priming} and classical conditioning \cite{hopkins2015eye,spataro2017implicit}. The task is to learn and store natural associations between stimuli and corresponding responses, thereby allowing appropriate responses to be invoked automatically in the right situations. This bypasses the need to perform complex cognitive computations required for reasoning, and, consequently, speeds up reaction times. The subconscious phenomena and how they interact with explicit representations such as facts are still poorly-understood.  Consequently, implementations of such techniques are lacking in the AI arena. 
	Procedural knowledge in large language models is commonly learned implicitly and stored in long-term memory. This is achieved by virtue of LLMs’ ability to learn sequences of actions, structures, and relationships in the training data. After the pre-training phase, LLMs can further be fine-tuned on task-specific, sensorimotor datasets (e.g., \cite{mon2024robotic} which usually include detailed step-by-step instructions on how to accomplish the target tasks. The fine-tuning process is aimed at internalizing (i.e., learning from scratch) or refining (i.e., align slightly different tasks) the model’s ability to generate and follow explicit execution plans necessary for solving the given problem. Despite the impressive performance of state-of-the-art models on procedural tasks, it is often more effective to encode specific skills explicitly with the help of neuro-symbolic techniques \cite{kwon2024fast,aktas2024vq,silver2024generalized}. These symbolic methods are sometimes used to provide structured reasoning frameworks that can be utilized by implicitly learned procedural knowledge for solving specific sets of problems.

	\section{Generalist AI (AGI) framework based on the principles of embodiment, grounding, causality and memory}
\begin{figure*}[!htb]
		\vspace {-1mm}
		\centering
		\includegraphics[width=1.0 \linewidth]{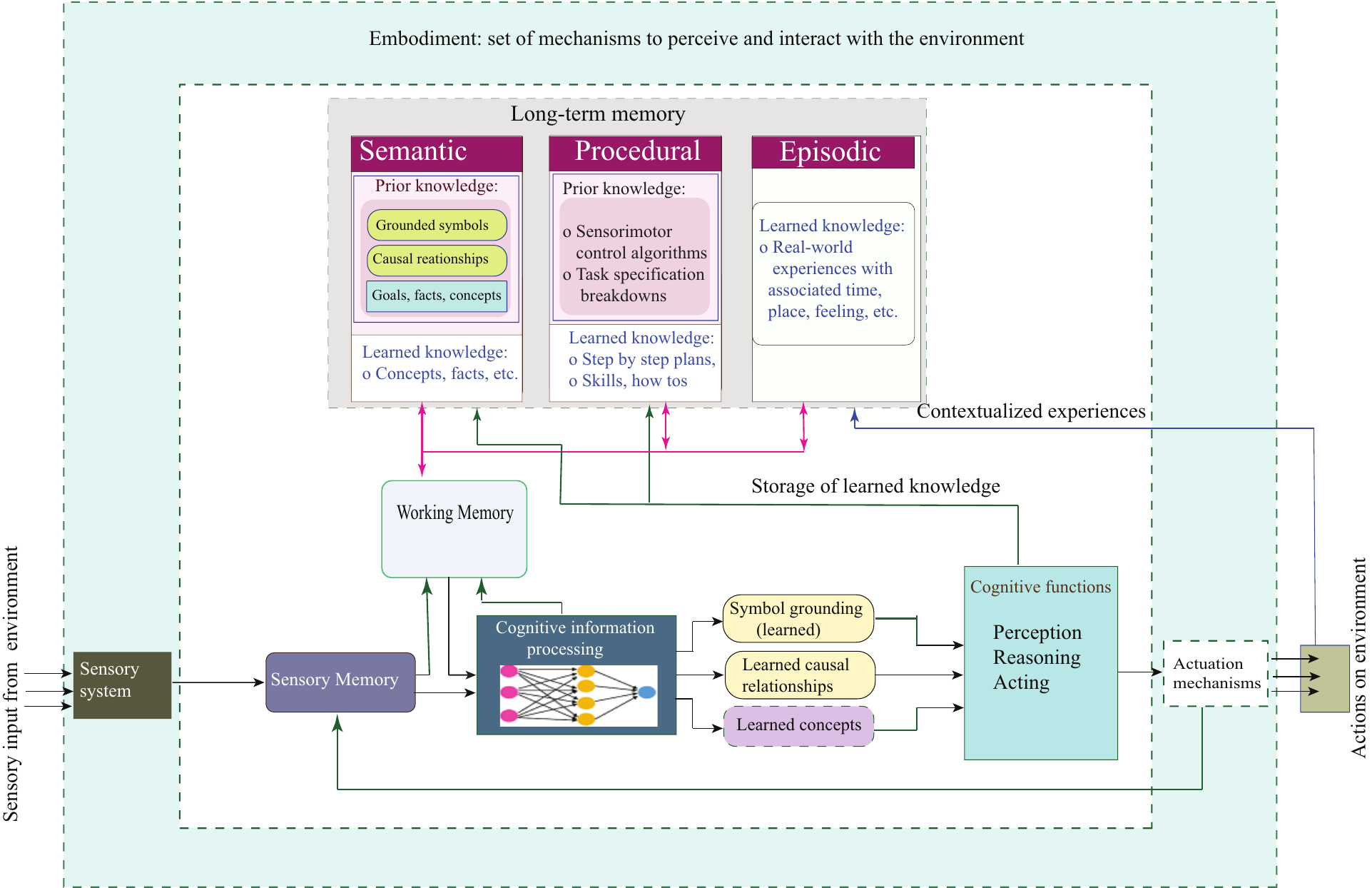} \vspace {-3mm}
		\caption{Functional block diagram of a generalized AGI system based on the principles covered in this article. The conceptual model consists of (1) a core framework, embodiment, that provides the physical essence and necessary mechanisms for interfacing and interacting with the world; (2) memory, made up of different memory subsystems – sensory, working and long-term memory, whose role, among others, is to allow both learned and prior knowledge to be preserved and accumulated over time; (3) symbol grounding subsystem, which provides a way to connect abstract representations in the underlying LLM model to actual entities in the world; and (4) causal learning mechanisms that learn the properties and physical laws associated with entities of the real world.  Note that the symbol grounding and causal learning mechanisms combine both prior knowledge encoded in memory and learned knowledge resulting from cognitive information processing to achieve correct results.   
		}\label{fig:X3_AGI_Framework}
	\end{figure*}

	In this section we develop a conceptual framework for AGI that unifies the concepts covered in this work. Such a framework implements the essential computational mechanisms that support the realization of sophisticated, robust and general intelligence based on the principles discussed in Sections 2-5 of this paper. These seemingly isolated concepts surveyed in this paper– embodiment, grounding, causality and memory – are inherently interrelated and complementary in their functions with regard to their role in facilitating artificial general intelligence. 
	Embodiment provides the general structure and requisite mechanisms for interfacing with the world. This allows AI systems to experience the world (through sensing systems) and influencing its state by performing desired actions in response to sensory inputs and goals. These embodied experiences serve as useful signals for grounding symbols. That is, rather than relying solely on abstract, linguistic associations of input words in training data, embodiment allows the agent to obtain meaningful sensorimotor experiences – through actual perception of and interactions with the world – that is used to ground abstract representations in the agent’s actual perception of and interactions. This grounded embodied experiences, in turn, allows the agent to observe and learn causal relationships directly from the physical world through interaction with and feedback from it. Additionally, memory mechanisms provide a means for encoding, storing and accessing grounded symbols, embodied experiences and causal relationships learned in the process of training the AI agent (see Figure \ref{fig:X3_AGI_Framework}).  Moreover, memory serves as a means to incorporate already-known causal relationships and grounded symbols as prior knowledge. As shown in Figure \ref{fig:X3_AGI_Framework}, the causal knowledge and symbolic associations learned from agent observations and embodied interactions with the environment can then be augmented with the prior known causal relationships and grounded symbols encoded in memory in the form of structured knowledge, providing a more comprehensive knowledge for robust perception, reasoning and other cognitive tasks.  These mechanisms, working together, form a robust framework that enables LLM agents to generalize knowledge better.


	Functional block diagram of a generalized AGI system based on the principles covered in this article. The conceptual model consists of (1) a core framework, embodiment, that provides the physical essence and necessary mechanisms for interfacing and interacting with the world; (2) memory, made up of different memory subsystems – sensory, working and long-term memory, whose role, among others, is to allow both learned and prior knowledge to be preserved and accumulated over time; (3) symbol grounding subsystem, which provides a way to connect abstract representations in the underlying LLM model to actual entities in the world; and (4) causal learning mechanisms that learn the properties and physical laws associated with entities of the real world.  Note that the symbol grounding and causal learning mechanisms combine both prior knowledge encoded in memory and learned knowledge resulting from cognitive information processing to achieve correct results.

	\section{Discussions}
	Large language models have surpassed traditional deep learning approaches in many tasks. They have achieved impressive results in many nontrivial AI problems, including reasoning, planning, multimedia (i.e., text, image, video, speech, etc.) generation, open-world navigation, coding, natural language understanding, and open-domain question answering. Because of these capabilities, commercial firms, including tech giants such as Google, OpenAI, Meta, Nvidia, Amazon, Apple and Microsoft have invested huge sums of money and human capital in developing generic as well as domains-specific generalist AI systems. State-of-the-art generalist AI systems are also increasingly being incorporated in commercial products such as search engines, chatbots, generic software, portable navigation equipment, smart phones, autonomous vehicles, and extended reality systems.
	The recent success of multimodal language models has drastically raised expectations about the possibility of machines achieving universal intelligence in the foreseeable future. In fact, some researchers are of the view that with state-of-the-art LLMs artificial general intelligence is already attainable \cite{y2023artificial,li2024artificial,ilic2024evidence}. While multimodal LLMs are showing enormous promise, these claims, at the moment, are premature and exaggerated. 
	A possible route to attaining artificial general intelligence is to continue to scale up large, universal machine learning algorithms and training them with yet larger and larger quantities of data so as to handle complex problems in a wide range of domains. Given the impressive results already achieved by state-of-the art neural network frameworks, especially multimodal large language models, such an approach seems possible, at least in principle. The large model size and the enormous volume and diversity of training data allows these models to capture general yet intricate concepts and semantically rich patterns and associations that are valid across multiple problem domains and application settings. However, experience shows that such an approach has serious limitations: limited data in many specialized domains, neural networks tend to learn mere data correlations and fail to distinguish between superficial associations and causal relationships.  Moreover, as demonstrated by state-of-the-art large language models (see, for example \cite{bender2021dangers,zevcevic2023causal,ray2023chatgpt,huang2023survey}), such intelligent systems will be very superficial with regard to the sophistication of its knowledge and ability to apply learning in a flexible, context-dependent manner in unseen situations. Thus, simply scaling up LLMs and training on similarly larger datasets may not be enough to achieve human-level intelligence.

	Large language models are still unable to match the robustness, flexibility, efficiency and the overall generalist capabilities of the biological cognitive systems. In contrast to large language models and AI systems, human intelligence is extremely rich and multifaceted. Humans are able to make accurate judgments about the properties and behavior of objects without direct measurements. To mitigate this shortcoming, many works aim to achieve general intelligence by engineering specialized properties that make biological intelligence so powerful, robust, data-efficient, versatile and adaptive.  In particular, the concepts discussed here – embodiment, grounding, causality and memory –would be extremely useful in achieving significant milestones. While the principles are promising, there is still significant room to improve the approaches used to implement each of these concepts. It is also important to emphasize that each of the concepts only solves specific cognitive problems related to the realization of general, human-level intelligence. To facilitate general intelligence, however, it will be much more beneficial to incorporate all these principles and approaches in a more integrated manner in a single cognitive framework.

	Thus, while the concepts of embodiment, symbol grounding, causality and memory have been long recognized as the foundation for artificial general intelligence, and have been widely employed to advance the state-of-the-art in LMMs, continued progress toward AGI will require fundamentally new paradigms for designing LLMs that implement all these principles in a unified fashion.  Such design philosophies will involve integrating deep learning models with neuro-symbolic techniques that leverage prior information to encode constrains and physical properties of the real world.  This approach requires these core concepts to be treated as a unified set of interrelated and complementary primitives that jointly model intelligent agents and their environment.   The cognitive process is then reduced to interfacing the various subcomponents and processing and exchanging information between them.  The processed cognitive information is then leveraged to understand specific event of interest, interact with the world, explain observations and account for counterfactuals. In this context, it will be extremely important to handle situations that are absent or are sparsely represented in training datasets.  
	
	Another major challenge with AGI research is that although the goal of achieving human-level, general intelligence seems well-defined, evaluating and ascertaining when this is achieved is a challenging problem. In particular, while comparisons of the intelligence of AI systems and humans are usually based on performance on specific sets of tasks, there exist fundamental differences in how human and machine intelligence are designed and function. These differences are reflected in their respective strengths and weaknesses. For instance, biological intelligence developed through evolution out of the necessity of survival of the agent itself (or its offspring) in dynamic and hostile environments. Human intelligence is fundamentally fuzzy and broad in scope, adaptable, and includes subjective aspects such as emotional, social, creative reasoning capabilities. In contrast, machine intelligence is typically designed and optimized to directly solve specific set of problems – no matter how general these problems may be. Owing to these important differences relating to their nature, design objectives and specific abilities, comparisons may yield misleading results. Consequently, even when an AI agent achieves general performance on complex tasks comparable to humans, it will still be difficult to characterize it as such. Moreover, Intelligence as an abstract concept is a continuous metric that encompasses multiple dimensions and measuring it in objective terms is not feasible. Consequently, it is not practical to even assess how close state-of-the art LLMs are to achieving AGI.

	Nevertheless, as the capabilities of intelligent agents continue to improve, at the point where one is no longer able to distinguish between the decisions and actions of AI agents and those of humans in a wide variety of complex (virtual or real-world) settings, one can safely conclude that we have attained human-level, general intelligence, even if in a limited sense. Currently, state-of-the-art LLM agents increasingly perform complex tasks in complex human-centric environments, and can assume leadership roles and provide expert guidance in specific open-world settings, where trust as well as professional, social, emotional relationships may develop between agents and humans as a result of their interactions. These agents are increasingly exhibiting fundamental human characteristics and capabilities, including the ability to understand the emotional state of humans; empathize; respond to unexpected, random events; help and request help; collaborate with humans and other agents to jointly solve problems; and engage in meaningful dialog with humans. At this stage, we are not very far from some form of what can be described as general intelligence.

	\section{Conclusion}
	In this work we present what we consider the core enablers of robust and sophisticated cognitive capabilities that AI models based on large language models can leverage to achieve artificial general intelligence – embodiment, symbol grounding, causality and memory. While these concepts are by no means the only principles necessary for the realization of general intelligence, they form the fundamental building blocks which are essential for any AI system to achieve general intelligence in dealing with the real world. Integrating these techniques in LLMs an intrinsic way will lead to a fundamentally new set of important characteristics that natively support AGI. The core building blocks and techniques for realizing these principles are already available, at least in rudimentary form. As our understanding of these principles and the techniques for implementing them continue to improve, the prospect of achieving human-level general intelligence in the foreseeable future is within reach.

	\appendices


	\ifCLASSOPTIONcompsoc
	\section*{Acknowledgments}
	\else
	\section*{Acknowledgment}
	\fi

	The authors would like to thank...

	\ifCLASSOPTIONcaptionsoff
	\newpage
	\fi

\end{document}